\documentclass{article}
\usepackage[preprint]{neurips_2026}
\usepackage[utf8]{inputenc} % allow utf-8 input
\usepackage[T1]{fontenc}    % use 8-bit T1 fonts
\usepackage{hyperref}       % hyperlinks
\usepackage{url}            % simple URL typesetting
\usepackage{booktabs}       % professional-quality tables
\usepackage{amsfonts}       % blackboard math symbols
\usepackage{nicefrac}       % compact symbols for 1/2, etc.
\usepackage{makecell}  
\usepackage{graphicx}  
\usepackage{array}     
\usepackage{graphicx}
\usepackage{microtype}    
\usepackage{siunitx}
\usepackage{xcolor}       
\usepackage{amsmath} 
\usepackage{multirow} 
\usepackage{enumitem}
\usepackage{xspace}

\usepackage{tabularx}
\usepackage{longtable}
\usepackage{float}
\usepackage{caption}
\usepackage{subcaption}
\usepackage{listings}
\usepackage{natbib}
\usepackage{colortbl}
\usepackage{tcolorbox}
\tcbuselibrary{skins,breakable,listings}

\definecolor{promptbg}{RGB}{245,245,245}
\definecolor{promptframe}{RGB}{180,180,180}
\definecolor{mechblue}{RGB}{30,80,160}
\definecolor{lightgray}{RGB}{240,240,240}

\newcommand{\mechsim}{\ourmethod}

\newcommand{\ourmethod}{MechSim\xspace} 

\usepackage{xcolor}

\title{Simulate, Reason, Decide: Scientific Reasoning with LLMs for Simulation-Driven Decision Making}

\author{
    Yuhan Yang \quad Ruipu Li \quad Alexander Rodríguez \\
    Computer Science and Engineering \\
    University of Michigan \\
    \texttt{\{yuhanyyh, liruipu, alrodri\}@umich.edu}
}

\begin{document}

\maketitle

\begin{abstract}
Scientific simulators are increasingly being integrated into LLM-driven systems for high-stakes simulation-driven decision-making. However, existing frameworks primarily use LLMs to generate, calibrate, or execute simulators, treating them as black-box interfaces rather than as structured mechanistic systems that can be reasoned about. As a result, current approaches lack the ability to identify, represent, and reason about the assumptions and mechanisms underlying simulator behavior, limiting transparency, auditability, and decision justification. We introduce MechSim, a mechanism-grounded neuro-symbolic reasoning framework for executable scientific simulators. Unlike prior neuro-symbolic approaches that primarily reason over static symbolic structures, MechSim enables LLM agents to reason about the mechanisms, assumptions, and execution behavior of scientific simulators. Our framework represents simulators through a shared structured schema capturing assumptions, variables, mechanism dependencies, and execution traces. On top of this representation, LLM agents operate as constrained reasoning engines that generate structured, evidence-grounded explanations linking simulator outcomes to their underlying mechanisms. We evaluate our approach across multiple high-stakes domains and show that it improves mechanism-level explanation quality, simulator analysis, and downstream decision-making reliability.
\end{abstract}

\section{Introduction}
LLM agents provide a natural interface for automating decision-making workflows that ingest multimodal information from diverse sources and generate recommendations~\cite{ma2025agent,xiao2024tradingagents}. More recently, LLM-driven systems have begun incorporating scientific simulators to support simulation-driven decision-making in high-stakes domains such as healthcare, finance, logistics, and public policy~\cite{kim2024mdagents,datta2026agentic,choukhmane2026ai}. Scientific simulators are particularly valuable in these settings because they enable the evaluation of interventions and counterfactual scenarios when real-world experimentation is expensive, risky, or ethically infeasible~\cite{marathe2013computational,banks2010discrete,carson2005introduction}.

Despite growing interest in LLM-integrated simulation systems, existing approaches primarily use LLMs to generate, calibrate, or execute simulation models~\citep{holt2025gsim,datta2026agentic,wahl2026probabilistic}. However, they provide limited support for reasoning about the simulators themselves. In particular, current frameworks lack structured representations that allow LLM agents to analyze simulator assumptions, mechanism-level dependencies, and execution behavior in a grounded and verifiable manner. Consequently, these systems remain limited to descriptive summaries and surface-level interpretations of simulator outputs, often producing explanations that appear plausible but are not grounded in simulator structure, execution traces, or scientific evidence. %~\citep{ji2023survey,huang2025survey}.

This limitation is particularly important because scientific simulators explicitly encode assumptions, variables, and mechanism-level dependencies that fundamentally shape their outputs. However, these assumptions are often embedded within specialized scientific formalisms, making it difficult to determine why particular outcomes arise and whether simulator behavior is aligned with the deployment context or evaluated policies. For example, during a public health crisis, an epidemic simulator may predict that school closures reduce peak infections, yet it may remain unclear whether this effect is primarily driven by reduced child-to-child transmission, mobility changes, or assumptions embedded in the contact-network structure. More importantly, the validity of these conclusions may depend on whether the simulator assumptions align with the real-world context, such as local social-contact patterns, behavioral responses, or intervention compliance. Without a framework capable of tracing outcomes back to simulator assumptions and mechanisms, simulation-driven recommendations become difficult to interpret, audit, and validate, all of which are critical for high-stakes decision-making~\cite{rudin2019stop,saltelli2020five,holmdahl2020wrong}. %ioannidis2022forecasting

To address these challenges, we propose MechSim, a neuro-symbolic framework for reasoning about scientific simulators in simulation-driven decision-making. MechSim combines structured simulator representations with LLM-based reasoning to enable grounded analysis of simulator behavior and outcomes. Our main contributions are threefold. 
\begin{enumerate}[leftmargin=*]
    \item We introduce a \textbf{mechanism-grounded reasoning framework} that enables LLM agents to analyze how and why simulator outcomes arise, rather than operating solely on surface-level outputs. Our framework represents simulators through a shared structured schema capturing assumptions, variables, mechanism-level dependencies, and execution traces, enabling LLM agents to reason over simulator structure in a grounded and constrained manner. 
    \item We introduce a \textbf{verifiable explanation framework} in which generated claims must be grounded in simulator structure, execution traces, or retrieved scientific evidence. This transforms explanation generation from post-hoc summarization into a process of constructing and validating mechanism-grounded hypotheses about simulator behavior.
    \item We conduct extensive experiments across three high-stakes domains, demonstrating that MechSim improves explanation quality, simulator analysis, and downstream decision-making reliability across multiple scenarios.
\end{enumerate}

\section{Problem Formulation}
\label{sec:problem-setup}
We study the problem of enabling LLM agents to guide simulation-driven decision-making. In this setting, LLM agents must analyze simulator behavior, assumptions, outputs, generate recommendations, and provide grounded justifications for real-world decision tasks. We consider a setting consisting of a decision task $\mathcal{T}$, a set (ensemble) of structurally heterogeneous scientific simulators $\mathcal{S} = \{s_1, s_2, \dots, s_n\}$, contextual information $\Omega$ describing the deployment environment, and a scientific corpus $\mathcal{D}$ containing scientific literature, technical reports, or policy guidelines.

The contextual information $\Omega$ captures properties of the real-world environment in which decisions are made, including population characteristics, institutional constraints, geographic conditions, time horizons, behavioral patterns, and cultural factors that influence whether simulator assumptions are valid for the target setting~\citep{uhrmacher2024context}. Importantly, simulators within $\mathcal{S}$ may differ substantially in their assumptions, internal mechanisms, and computational abstractions, potentially producing divergent outputs under the same deployment context~\cite{holmdahl2020wrong}. As a result, simulator outputs alone are often insufficient for reliable decision-making, since recommendations may depend critically on latent simulator assumptions and their alignment with the deployment environment.

The task specification $\mathcal{T}$ defines the intended use of simulator outputs for decision-making. In this work, we focus on two primary scenarios. (1) \textbf{Policy Selection}: the agent evaluates candidate interventions based on simulator outputs, such as public-health interventions in epidemiology or inventory-management strategies in supply-chain systems. In this setting, $\mathcal{T}$ includes the set of candidate interventions to be evaluated. (2) \textbf{Simulator Selection for Forecasting}: prior to making predictions about future events, such as hospitalization forecasts or demand forecasts, the agent selects the most reliable simulator(s) from a set of available models.

Each simulator $s_i \in \mathcal{S}$ maps an initial condition $x$ to an output $y_i$ according to
$y_i = s_i(x, \theta_i, f_i)$,
where $\theta_i$ denotes simulator parameters and $f_i$ denotes the simulator's computational logic. In many scientific simulators, $x$ and $\theta_i$ are themselves uncertain and are therefore modeled as probability distributions rather than fixed values, an assumption we adopt in this work. The initial conditions $x$ and parameters $\theta_i$ are instantiated from the contextual information $\Omega$ and the task specification $\mathcal{T}$. %The central challenge is that simulator outputs alone do not reveal which assumptions, mechanisms, or contextual factors drive the resulting predictions. 
Our goal is therefore to enable LLM agents to generate recommendations and explanations that are grounded in simulator mechanisms, execution behavior, and contextual validity.

\section{MechSim: Mechanism-Aware Reasoning for Scientific Simulators}
\label{sec:method}

MechSim is a neuro-symbolic framework for mechanism-grounded constrained reasoning over executable simulators. The key challenge in our problem setting is that simulator outputs alone do not expose which assumptions, mechanisms, or contextual factors drive the resulting predictions. MechSim addresses this challenge by explicitly representing simulator structure and constraining LLM reasoning using simulator mechanisms, scientific evidence, and execution behavior, defined as the evolution of simulator states and mechanism activations during simulation.

Our framework decomposes simulator reasoning into four tightly coupled components: (1) contextual grounding, (2) mechanism-level grounding, (3) constrained mechanism-level reasoning, and (4) verification checks. The symbolic layer of MechSim consists of structured simulator representations, including mechanism graphs, assumptions, execution traces, and sensitivity-analysis artifacts. The neural layer consists of LLM-based contextual interpretation, scientific evidence synthesis, explanation generation, and decision reasoning. Mechanism graphs serve as the symbolic substrate over which neural reasoning operates, enabling grounded reasoning over executable simulators rather than unconstrained reasoning over textual summaries alone.

\begin{figure}[t]
    \centering
    \includegraphics[width=1\linewidth]{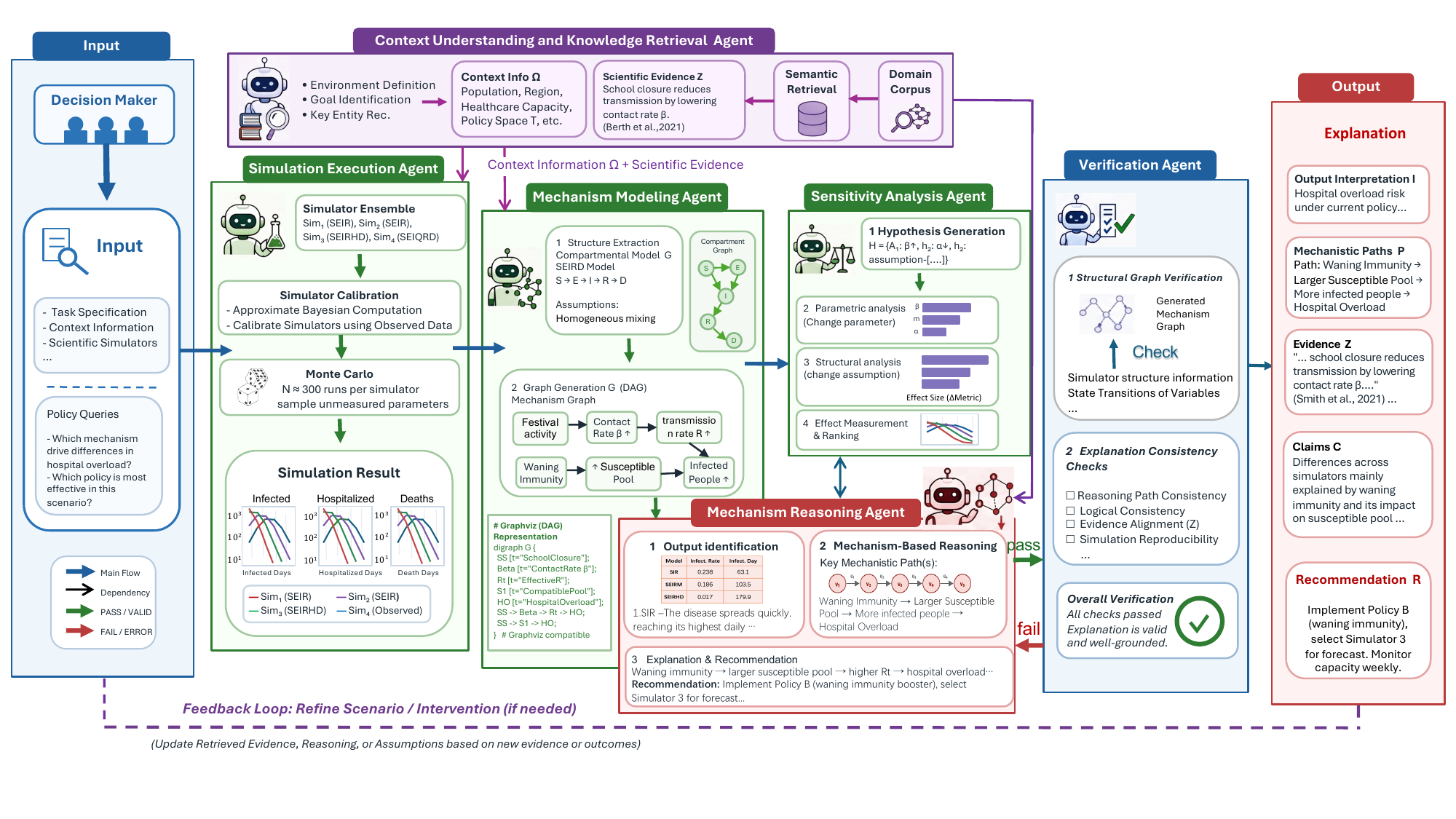}
    \vspace{-10mm}
    \caption{Overview of MechSim, a neuro-symbolic framework for mechanism-grounded reasoning over scientific simulators in simulation-driven decision-making. MechSim coordinates specialized LLM agents for contextual grounding, simulator execution, mechanism modeling, sensitivity analysis, constrained reasoning, and verification to generate empirically grounded explanations and decision recommendations from simulator outputs, mechanism graphs, and scientific evidence.}
    \label{fig:MechSim-framework}
\end{figure}

Each component of MechSim is instantiated through specialized modules implemented as cooperative LLM agents, as illustrated in Figure~\ref{fig:MechSim-framework}. These agents are responsible for contextual understanding, simulator calibration and execution, mechanism modeling, constrained reasoning, sensitivity analysis, and verification. While the framework is conceptually organized around grounded neuro-symbolic reasoning, the agent decomposition enables modular execution and iterative refinement across stages.

Formally, given contextual information $\Omega$, task specification $\mathcal{T}$, simulator ensemble $\mathcal{S}$, and scientific corpus $\mathcal{D}$, MechSim constructs structured simulator representations and generates a verified explanation $E$ and recommendation $R$
% :
% \begin{align}
%     (\mathcal{Z}, \mathcal{Y}, \mathcal{G}) &= \textsc{MechSim-Construct}(\Omega, \mathcal{T}, \mathcal{S}, \mathcal{D}), \\
%     E, R &= \textsc{MechSim-Reason}(\Omega, \mathcal{T}, \mathcal{S}, \mathcal{Z}, \mathcal{Y}, \mathcal{G}),
% \end{align}
subject to structural consistency, evidence grounding, and empirical support constraints.
It is noteworthy that unlike prior neuro-symbolic reasoning approaches that primarily operate over static symbolic structures, MechSim performs constrained reasoning over executable simulators whose behavior emerges dynamically through simulation and intervention.

\subsection{Contextual Grounding}

The contextual grounding component, implemented through a Context Understanding Agent, transforms heterogeneous contextual information into structured representations suitable for downstream reasoning. Given contextual information $\Omega$, task specification $\mathcal{T}$, and scientific corpus $\mathcal{D}$, MechSim extracts decision-relevant entities, constraints, interventions, and initial conditions required for simulator execution and reasoning. In parallel, the framework retrieves scientific evidence $\mathcal{Z}$ to ground downstream explanation generation and verification. Specifically, given a query $q$ derived from $\Omega$ and $\mathcal{T}$, the retrieval function
$\mathcal{K}(q,\mathcal{D}) \rightarrow \mathcal{Z}$
retrieves relevant scientific evidence using a dense retrieval and reranking pipeline inspired by OpenScholar~\citep{asai2026synthesizing}. The resulting evidence set $\mathcal{Z}$ provides scientific grounding for downstream reasoning and reduces unsupported or hallucinated explanations~\citep{ji2023survey,huang2025survey}.

\subsection{Mechanism-Level Grounding}

Mechanism-level grounding constructs structured representations of simulator behavior and execution dynamics. The central goal is to expose the mechanisms, assumptions, and execution patterns that produce simulator outcomes, enabling grounded reasoning over simulator behavior rather than reasoning solely over outputs.

\paragraph{Mechanism Graph Construction.}

For each simulator $s_i \in \mathcal{S}$, MechSim constructs a mechanism graph
$\mathcal{G}_i = (\mathcal{V}_i, \mathcal{E}_i, \mathcal{M}_i)$,
where $\mathcal{V}_i$ denotes simulator variables, $\mathcal{E}_i$ denotes mechanism-level transitions between variables, and $\mathcal{M}_i$ denotes graph metadata associated with the simulator. The metadata $\mathcal{M}_i$ includes simulator assumptions $\mathcal{A}_i$, execution traces $\tau_i$, and summary statistics derived from simulator execution.
The mechanism graph serves as the symbolic representation used to constrain downstream reasoning. Nodes correspond to simulator entities or state variables, while edges represent mechanistic transitions, interactions, or propagation pathways encoded by the simulator. Assumptions $\mathcal{A}_i$ determine which mechanisms are active under a given deployment context and intervention setting.

To construct $\mathcal{G}_i$, the Mechanism Modeling Agent extracts structured simulator representations from simulator definitions, execution traces, and domain documentation. Simulation outputs such as peak infection, mortality, or system overload are attached as graph-level metadata, enabling joint reasoning over simulator structure and observed outcomes.

\paragraph{Simulator Calibration and Execution.}

Given contextual information $\Omega$ and task specification $\mathcal{T}$, MechSim instantiates and executes each simulator in the ensemble $\mathcal{S}$. Initial conditions and intervention settings are derived from the deployment context, observational data, and task constraints.

Because simulator parameters are often uncertain, we model them as probability distributions rather than fixed values. We use approximate Bayesian computation (ABC)~\citep{sisson2018handbook,temfack2025sequential} to estimate posterior parameter distributions (calibration) for each simulator, enabling uncertainty-aware execution and robust comparison across simulators.
The Simulation Execution Agent runs each simulator and produces outputs 
$\mathcal{Y} = \{y_i\}_{i=1}^n$, along with execution traces (observable dynamical trajectories and intermediate states generated during simulation) and summary statistics used for downstream reasoning and verification. These artifacts provide empirical grounding for mechanism-level reasoning. 

\paragraph{Sensitivity Analysis as Executable Verification.}
\label{sec:sensitivity}

Sensitivity analysis operationalizes executable hypothesis testing over simulator mechanisms. Rather than treating explanations as purely linguistic artifacts, our Sensitivity Analysis Agent evaluates whether proposed explanations are empirically supported through controlled interventions on simulator parameters and assumptions~\citep{saltelli2008global}.

For parametric sensitivity analysis, numerical parameters $\theta_i$ are perturbed to identify variables that most strongly influence simulator outputs. For structural sensitivity analysis, simulator assumptions $\mathcal{A}_i$ and mechanism components within $\mathcal{G}_i$ are modified to evaluate how changes in simulator structure alter observed outcomes. To quantify these effects, MechSim re-executes simulator $S_i$ under alternative configurations of $\phi$:
\begin{equation}
    \text{Sensitivity}(\phi) =
    \mathbb{E}_{s_i(\phi=\phi_2)}[y]
    -
    \mathbb{E}_{s_i(\phi=\phi_1)}[y],
\end{equation}
where $\phi \in \{\theta_i, \mathcal{A}_i\}$ and $\phi_1,\phi_2$ denote alternative parameter settings or assumptions. These interventions provide empirical evidence supporting, rejecting, or refining candidate explanations.

\subsection{Constrained Mechanism-Level Reasoning}

Our Mechanism Reasoning Agent performs constrained reasoning over mechanism graphs, simulator outputs, execution traces, sensitivity-analysis artifacts, and retrieved scientific evidence. Unlike unconstrained LLM reasoning, the framework restricts explanation generation to reasoning paths supported by simulator structure, execution behavior, and empirical evidence.

Given mechanism graphs $\mathcal{G}$, simulator outputs $\mathcal{Y}$, sensitivity-analysis results, and retrieved evidence $\mathcal{Z}$, we construct explanations by tracing how simulator assumptions activate mechanism-level propagation paths that produce decision-relevant outcomes. Rather than relying on surface-level metric comparison, the framework reasons over node-to-node transitions within mechanism graphs and explicitly links generated claims to simulator mechanisms, assumptions, evidence, and execution behavior. Importantly, when evidence is incomplete, conflicting, or insufficient to support a claim, the framework explicitly represents the corresponding uncertainty in its explanations. 

Drawing on foundational work in explanation and argumentative reasoning (Toulmin model)~\citep{miller2019explanation,toulmin2003uses}, MechSim constructs a structured explanation:
    $E = (\mathcal{I}, \mathcal{P}, \mathcal{Z}, \mathcal{C})$,
where:
\begin{enumerate}[leftmargin=*]
    \item \textbf{Output Interpretation ($\mathcal{I}$):} synthesizes the context, task, and simulator outputs, identifying decision-relevant output patterns for downstream reasoning and decision-making. %summarizes decision-relevant simulator behavior and output patterns conditioned on the task and deployment context.
    \item \textbf{Reasoning Paths ($\mathcal{P}$):} $\mathcal{P} = \{p_1, p_2, \dots, p_k\}$ denotes a set of mechanism propagation paths under a simulator-specific assumption set $\mathcal{A}_i$. Each path $p \mid \mathcal{A}_i = (v_1 \xrightarrow{e_1} v_2 \xrightarrow{e_2} \dots \xrightarrow{e_{n-1}} v_n)$ represents an ordered sequence of variables connected by mechanisms activated under $\mathcal{A}_i$, showing how assumptions and mechanisms lead to outcomes relevant to the task~\citep{machamer2000thinking}. 
    \item \textbf{Supporting Evidence ($\mathcal{Z}$):} contains scientific evidence retrieved from the domain corpus $\mathcal{D}$. The evidence provides scientific grounding and is explicitly referenced to support the claims in $\mathcal{C}$~\citep{xu2025simulrag}. %contains retrieved scientific evidence used to ground reasoning and validate generated claims.
    \item \textbf{Claims ($\mathcal{C}$):} $\mathcal{C} = \{c_1, c_2, \dots, c_m\}$ provides mechanism-grounded claims about the simulator results. Each claim integrates mechanism paths and external evidence to translate simulator outcomes into interpretable explanations that justify decisions and recommendations. %contains mechanism-grounded claims connecting simulator behavior to decision-relevant recommendations and interpretations.
\end{enumerate}

\subsection{Verification Checks}

The final stage is implemented by a Verification Agent that verifies that explanations and recommendations are faithful to simulator mechanisms, execution behavior, scientific evidence, and empirical validation results. The goal is to ensure that generated explanations are not merely linguistically plausible but are structurally and empirically grounded~\citep{sargent2010verification}.

A valid explanation $E=(\mathcal{I},\mathcal{P},\mathcal{Z},\mathcal{C})$ and recommendation $R$ must satisfy three properties: (1) \textbf{Structural Consistency:} each claim $c \in \mathcal{C}$ must be traceable to valid reasoning paths within mechanism graphs $\mathcal{G}_i$; (2) \textbf{Evidence Grounding:} generated claims must be logically consistent with retrieved scientific evidence $\mathcal{Z}$ and simulator outputs $\mathcal{Y}$; (3) \textbf{Empirical Support:} reasoning paths and claims must be supported by executable verification through sensitivity analysis when available.

If verification detects unsupported claims, invalid reasoning paths, or inconsistencies between explanations and simulator behavior, the framework initiates iterative refinement before producing final outputs. After verification, MechSim produces a verified explanation $E$ and decision recommendation $R$. For policy selection, recommendations may identify interventions supported by simulator evidence and contextual assumptions. For simulator selection for forecasting, recommendations may identify which simulator is most reliable under the target deployment context. Importantly, recommendations are generated jointly with explanations and finalized only after verification.

\section{Experiments and Evaluation}
We evaluate \ourmethod across three high-stakes domains and three tasks: explanation generation, policy selection, and simulator selection for forecasting. We compare \ourmethod with representative LLM-based reasoning baselines. 
 
\subsection{Benchmark Environments}
We use three simulation environments constructed from, or inspired by, real-world scenarios. Each environment includes an oracle simulator that generates ground-truth outcomes under different policies, along with a set of candidate simulators for modeling the dynamics. For each environment, we run 100 simulations by sampling the oracle simulator parameters and initial conditions from a predefined distribution and report the aggregated metrics. 

(1) \textbf{\textit{COVID-19}}: We study COVID-19 transmission under alternative intervention policies using six compartmental epidemic simulators: SIR, SEIRD, SEIRHD, SEIQRD, SEAIHRD, and SIRV. This environment is motivated by COVID-19 scenario modeling, where compartmental simulators are calibrated to observed data and used to project epidemic trajectories under different interventions~\cite{howerton2023evaluation}. We use a stochastic, multi-compartment, discrete-time simulator as the oracle simulator and ground truth~\cite{kain2021chopping}. For the forecast task, we use real-world COVID-19 data, including daily confirmed cases and deaths aggregated at the US state and national level~\cite{cramer2022united}.

(2) \textbf{\textit{Supply Chain}}: We analyze inventory management and demand fluctuations in multi-tier logistics systems using four candidate simulators: base-stock inventory control, $(s,S)$ inventory control\footnote{Here, $s$ refers to the reorder point in the inventory policy, which is different from the $s$ used to denote simulators  earlier.}, decentralized multi-echelon $(r,Q)$ control, and the Beer Game supply-chain model. The oracle simulator is a stochastic multi-echelon supply-chain simulator with intervention policies that alter managerial levers such as safety stock, lead times, demand sharing, order quantities, and expedited replenishment \cite{sterman2000business}. For the dataset, we use the M5 dataset, which includes hierarchical point-of-sale records at the item–store level in Walmart \cite{makridakis2022m5}.

(3) \textbf{\textit{Measles}}: We study measles transmission and vaccination dynamics across six configured outbreak scenarios, including three source-informed configurations based on London, Chicago, and Texas--New Mexico outbreaks \cite{allotey2017comparison, masters2024real, gonzalez2025modeling}, and three synthetic configurations that simulate outbreaks in small populations such as schools. We use compartmental mechanistic baseline simulators implemented with Epydemix, including SIR, SEIR, SIRV, SVEIR, and SVVIR models~\cite{kuddus2021mathematical}. 
The oracle simulator is a stochastic agent-based \texttt{measles:ModelMeaslesMixing} model accessed through \texttt{rpy2}/epiworldR~\cite{epiworldR2023,R-measles}. %,R-epiworldR,R-measles,measles2023
Although measles and COVID-19 are both respiratory viral infections, they differ substantially in transmission dynamics and intervention strategies~\cite{gaythorpe2021impact}. As a result, their simulators rely on distinct assumptions and encode different mechanisms.

\subsection{Baseline Methods}
We compare \ourmethod with three baselines. For a fair comparison, all baselines reason over the same simulation outputs as \ourmethod and receive the same task information. (1) \textbf{\textit{Causal-Copilot}} \cite{wang2025causal} is an autonomous agent that operationalizes expert-level causal analysis within an LLM framework. (2) \textbf{\textit{Logic-LM}} \cite{pan2023logic} integrates symbolic reasoning with LLMs to support structured, logically grounded analysis. (3) \textbf{\textit{Graph of Thoughts (GoT)}} \cite{besta2024graph} organizes inference as a graph, allowing the model to explore and aggregate multiple reasoning paths. These baselines were selected because they represent three distinct categories of reasoning methods and are both representative and reproducible.
More information about baseline methods in available is Appendix \ref{app:baselines}.

\subsection{Tasks}

\par\textbf{Policy Selection Task}: This task mimics the real-world decision-making process of selecting an optimal policy. For each simulation environment, we use a sophisticated, realistic simulator as the oracle model. The oracle model first generates ground-truth observations, then generates that under different policies, allowing us to evaluate the effect of policies and obtain the ground-truth policy ranking. In each experiment, we test 6–9 candidate policies, with specific details provided in the Appendix \ref{app:policies}. To test the robustness of our method, we sampled the parameters across 100 scenarios and recorded the average metrics. Each method then recommends a ranking of the candidate policies, and evaluation metrics are averaged over domain-specific sampled scenarios or stochastic replicates, as detailed in the Appendix \ref{app:experiments}. 

\textit{Evaluation Metrics.} In the main results, we report Precision@3, Precision@5, Recall@3, and Recall@5 \cite{manning2008introduction,liu2009learning}. These ranking metrics compare each method's top-$K$ recommended policies with the oracle-preferred policy set: Precision@$K$ measures the fraction of recommended top-$K$ policies that are oracle-preferred, while Recall@$K$ measures how much of the oracle-preferred set is recovered by the method's top-$K$ recommendations. Higher precision and recall indicate stronger alignment with the oracle policy ranking.

\par\textbf{Simulator Selection for Forecasting Task}: This task requires selecting the most suitable simulator for forecasting under varying conditions. In practice, a decision-maker may have access to multiple calibrated simulators, each fit to real-world data but differing in its assumptions, mechanisms, and predictive behavior. During the pre-cutoff period, each simulator generates forecasts, and its predictive error is measured using Root Mean Square Error (RMSE). Each method then recommends a ranking of the candidate simulators given the pre-cutoff observations. The evaluation metrics are averaged over domain-specific forecast instances. 
Each domain evaluates simulator rankings on held-out forecasts: COVID-19 uses 4-week-ahead rolling forecasts over weekly death data across a 60-week held-out period, Supply Chain uses a 28-day demand forecast window, and Measles uses the first 14 days of incidence for fitting before forecasting the remaining held-out horizon (around 1 year).

\textit{Evaluation Metrics.}
We compute \textit{regret} to evaluate the recommended simulators using ground-truth observations from the post-cutoff period. Regret is the gap between the RMSE of the selected simulator and that of the optimal simulator, i.e., the simulator with the lowest post-hoc RMSE \cite{cesa2006prediction}:
$\text{Regret} = \text{RMSE}(\hat{s}) - \min_{s \in \mathcal{S}} \text{RMSE}(s),$
where $\hat{s}$ denotes the selected simulator and $\mathcal{S}$ is the set of candidate simulators. For more than one selected simulators, we compute the average of the regret.
Lower regret indicates better performance.

\par\textbf{Explanation Evaluation Task}: This task evaluates whether a method can provide reliable explanations of simulator behavior in realistic decision-making scenarios. Given outputs from multiple simulators, each method explains the mechanisms that drive the observed outcomes. All evaluations are performed over five rounds with different random seeds, and the metrics are averaged.

\textit{Evaluation Metrics.} To evaluate the explanations, we use multiple LLMs as independent evaluators and conduct multi-round scoring based on four criteria: completeness, mechanistic consistency, scientific soundness, and faithfulness, inspired by prior work~\cite{deyoung2020eraser,liu2023g}.

\section{Main Results}
\label{sec:results}

We report the performance of our framework in three evaluation tasks across three benchmark domains: COVID-19, Supply Chain, and Measles.

\subsection{Policy Selection Results}
Table~\ref{tab:policy_selection} evaluates policy selection as a ranked decision problem, using Precision@$k$ and Recall@$k$ to measure whether each method identifies the oracle-preferred policies. \ourmethod obtains the best result in almost all settings, with especially consistent gains on COVID-19 and Supply Chain. This pattern is stable across LLM backbones: \ourmethod is best in $11/12$ Claude Sonnet 4.6 entries, $10/12$ DeepSeek V3.2 entries, and $10/12$ Qwen 3-235B entries. Compared with baselines emphasizing causal analysis~\citep{wang2025causal}, symbolic reasoning~\citep{pan2023logic}, and graph-structured deliberation~\citep{besta2024graph}, \ourmethod combines  mechanism abstraction, empirical sensitivity analysis, retrieval, reasoning, and verification. The policy-selection results suggest a benefit from the framework's integration of these components. 

On Measles, while \ourmethod excels in P@5 and R@5, Logic-LM or GoT is competitive on several P@3 or R@3 entries. This variability may reflect the greater heterogeneity of the Measles outbreak configurations and a smaller population size in these scenarios. These factors can make the oracle-preferred policy ranking less stable across top-$k$ metrics. Full results with standard deviations are in Appendix~\ref{com_policy_selection}.

\begin{table*}[t]
\centering
\scriptsize
\setlength{\tabcolsep}{3pt}
\renewcommand{\arraystretch}{1.15}
\caption{Policy selection results (Precision@$k$ and Recall@$k$) across three domains and three LLM backbones. \textbf{Bold} = best per column within each domain. “Causal” denotes Causal-Copilot, "Logic" refers to Logic-LM, and “GoT” denotes Graph of Thoughts.}
\label{tab:policy_selection}
\begin{tabular*}{\textwidth}{@{\extracolsep{\fill}} l l cccc cccc cccc @{}}
\toprule
& & \multicolumn{4}{c}{\textbf{Claude Sonnet 4.6}} & \multicolumn{4}{c}{\textbf{DeepSeek V3.2}} & \multicolumn{4}{c}{\textbf{Qwen 3-235B}} \\
\cmidrule(lr){3-6}\cmidrule(lr){7-10}\cmidrule(lr){11-14}
\textbf{Domain} & \textbf{Method} & P@3 & P@5 & R@3 & R@5 & P@3 & P@5 & R@3 & R@5 & P@3 & P@5 & R@3 & R@5 \\
\midrule
\multirow{4}{*}{COVID-19}
 & Causal. & 0.67 & 0.65 & 0.40 & 0.70 & 0.63 & \textbf{0.72} & 0.38 & 0.70 & 0.67 & 0.68 & 0.64 & 0.58 \\
 & Logic   & 0.63 & 0.62 & 0.38 & 0.65 & 0.60 & 0.61 & 0.36 & 0.64 & 0.70 & 0.60 & 0.42 & 0.67 \\
 & GoT        & 0.71 & 0.60 & 0.42 & 0.62 & 0.67 & 0.59 & 0.42 & 0.52 & 0.80 & 0.52 & 0.65 & 0.72 \\
 & MechSim    & \textbf{0.82} & \textbf{0.72} & \textbf{0.54} & \textbf{0.72} & \textbf{0.83} & 0.68 & \textbf{0.50} & \textbf{0.72} & \textbf{0.83} & \textbf{0.70} & \textbf{0.76} & \textbf{0.77} \\
\midrule
\multirow{4}{*}{\makecell[l]{Supply\\Chain}}
 & Causal. & 0.77 & 0.79 & 0.47 & 0.54 & 0.91 & 0.79 & 0.57 & 0.54 & 0.86 & 0.83 & 0.51 & 0.49 \\
 & Logic   & 0.82 & 0.83 & 0.49 & 0.35 & 0.87 & 0.85 & 0.52 & 0.65 & 0.87 & \textbf{0.86} & 0.52 & 0.49 \\
 & GoT        & 0.80 & 0.74 & 0.51 & 0.49 & 0.83 & 0.61 & 0.50 & 0.61 & 0.85 & 0.84 & 0.51 & 0.50 \\
 & MechSim    & \textbf{0.90} & \textbf{0.84} & \textbf{0.54} & \textbf{0.57} & \textbf{0.97} & \textbf{0.86} & \textbf{0.58} & \textbf{0.66} & \textbf{0.91} & 0.84 & \textbf{0.55} & \textbf{0.52} \\
\midrule
\multirow{4}{*}{Measles}
 & Causal. & 0.56 & 0.73 & 0.36 & 0.74 & 0.61 & 0.77 & 0.37 & 0.77 & 0.67 & 0.72 & 0.40 & 0.79 \\
 & Logic   & \textbf{0.72} & 0.77 & 0.37 & 0.79 & 0.69 & 0.67 & \textbf{0.43} & 0.65 & 0.72 & 0.73 & 0.47 & 0.73 \\
 & GoT        & 0.33 & 0.63 & 0.29 & 0.66 & 0.58 & 0.43 & 0.43 & 0.65 & 0.33 & 0.65 & \textbf{0.58} & 0.53 \\
 & MechSim    & 0.61 & \textbf{0.83} & \textbf{0.43} & \textbf{0.83} & \textbf{0.72} & \textbf{0.78} & 0.38 & \textbf{0.79} & \textbf{0.78} & \textbf{0.83} & 0.43 & \textbf{0.81} \\
\bottomrule
\end{tabular*}
\end{table*}

\subsection{Simulator Selection for Forecasting Results}
Table~\ref{tab:simulator_selection} evaluates simulator selection for forecasting using Top-$k$ regret, where lower regret means that the recommended simulator is closer to the post-cutoff RMSE-optimal simulator. \ourmethod obtains the lowest or tied-lowest regret in almost all settings. The gains are especially pronounced in Supply Chain, where \ourmethod reduces regret relative to the best baseline by $21.6\%$ for Claude Sonnet 4.6 Top-1, $20.8\%$ for Claude Sonnet 4.6 Top-3, and $59.3\%$ for Qwen 3-235B Top-1. COVID-19 shows smaller but consistent improvements for Top-1 regret across all three LLM backbones, consistent with the intended role of mechanism-grounded reasoning in identifying which simulator best captures the relevant dynamics. In the two settings where \ourmethod is not best, it remains close to the best score in absolute regret: it trails Causal-Copilot by $0.01$ on COVID-19 DeepSeek V3.2 Top-3 regret and by $0.15$ on Measles DeepSeek V3.2 Top-1 regret. In addition, similar to the policy selection results, \ourmethod yields consistent gains across LLM backbones, demonstrating that the framework's benefits are not specific to a particular LLM. 
Full results with standard deviations are in Appendix~\ref{app_simulator_selection}.

\begin{table*}[t]
\centering
\scriptsize
\setlength{\tabcolsep}{3pt}
\renewcommand{\arraystretch}{1.15}
\caption{Simulator selection results for the forecasting task: Top-$k$ regret ($k\in\{1,3\}$; \emph{lower is better}) across three domains and three LLM backbones. \textbf{Bold} denotes the best (lowest) result per row within each LLM block. “Causal” denotes Causal-Copilot, "Logic" refers to Logic-LM, and “GoT” denotes Graph of Thoughts.}
\label{tab:simulator_selection}
\begin{tabular*}{\textwidth}{@{\extracolsep{\fill}} l l cccc cccc cccc @{}}
\toprule
& & \multicolumn{4}{c}{\textbf{Claude Sonnet 4.6}} & \multicolumn{4}{c}{\textbf{DeepSeek V3.2}} & \multicolumn{4}{c}{\textbf{Qwen3-235B}} \\
\cmidrule(lr){3-6}\cmidrule(lr){7-10}\cmidrule(lr){11-14}
\textbf{Domain} & \textbf{Metric} & Causal & Logic & GoT & MechSim & Causal & Logic & GoT & MechSim & Causal & Logic & GoT & MechSim \\
\midrule
\multirow{2}{*}{COVID-19}
 & Top-1 Regret & 2.59 & 3.91 & 3.80 & \textbf{2.04} & 2.38 & 2.70 & 2.47 & \textbf{2.16} & 1.79 & 1.78 & 1.76 & \textbf{1.65} \\
 & Top-3 Regret & 2.60 & 2.41 & 2.53 & \textbf{2.29} & \textbf{2.10} & 2.14 & 2.18 & 2.11 & 1.70 & 1.85 & 1.70 & \textbf{1.68} \\
\midrule
\multirow{2}{*}{\makecell[l]{Supply\\Chain}}
 & Top-1 Regret & 6.17 & 5.42 & 5.92 & \textbf{4.25} & 0.62 & 0.54 & 0.59 & \textbf{0.47} & 0.59 & 0.62 & 0.59 & \textbf{0.24} \\
 & Top-3 Regret & 5.35 & 4.38 & 5.14 & \textbf{3.47} & 0.54 & 0.44 & 0.51 & \textbf{0.39} & 0.56 & 0.51 & 0.56 & \textbf{0.51} \\
\midrule
\multirow{2}{*}{Measles}
 & Top-1 Regret & 2.88 & 2.35 & 1.93 & \textbf{1.69} & \textbf{0.55} & 5.94 & 2.49 & 0.70 & 1.87 & 2.69 & 3.04 & \textbf{1.13} \\
 & Top-3 Regret & 3.40 & 3.21 & 2.79 & \textbf{2.20} & 2.99 & 3.26 & 2.76 & \textbf{2.28} & 2.55 & 3.17 & 2.70 & \textbf{2.49} \\
\bottomrule
\end{tabular*}
\end{table*}

\subsection{Explanation Evaluation Results}
Table~\ref{tab:main_comparison} evaluates explanation quality using a 1--5 judge score across four dimensions: completeness, mechanism-logic consistency, scientific soundness, and faithfulness. \ourmethod obtains the best score in $31/36$ cells. The gains are most consistent in Supply Chain, where \ourmethod ranks first on every criterion for all three LLM backbones. In COVID-19, \ourmethod remains competitive but is usually second on completeness and faithfulness.
\begin{table*}[t]
\centering
\small
\setlength{\tabcolsep}{4pt}
\renewcommand{\arraystretch}{1.15}

\caption{Comparison of different LLMs and reasoning methods across domains.
Metrics include Completeness (Comp.), Mechanism Logic Consistency (Logic.),
Scientific Soundness (Sci.), and Faithfulness (Faith.). “Causal” denotes Causal-Copilot, "Logic" refers to Logic-LM, and “GoT” denotes Graph of Thoughts.}
\label{tab:main_comparison}

\resizebox{\textwidth}{!}{%

\begin{tabular}{ll|cccc|cccc|cccc}
\toprule
\textbf{LLM} & \textbf{Method}
& \multicolumn{4}{c|}{\textbf{COVID-19}}
& \multicolumn{4}{c|}{\textbf{Supply Chain}}
& \multicolumn{4}{c}{\textbf{Measles}} \\
\cmidrule(lr){3-6} \cmidrule(lr){7-10} \cmidrule(lr){11-14}

&
& Comp. & Logic. & Sci. & Faith.
& Comp. & Logic. & Sci. & Faith.
& Comp. & Logic. & Sci. & Faith. \\
\midrule

\multirow{4}{*}{Claude Sonnet 4.6}
& Causal & 2.2 & 2.6 & 2.4 & 2.8 & 2.2 & 2.2 & 2.2 & 2.8 & 2.8 & 3.2 & 2.2 & 2.4 \\
& Logic       & 3.4 & 3.4 & 3.4 & 3.4 & 2.2 & 1.6 & 1.6 & 1.8 & 3.6 & 3.4 & 3.4 & 4.4 \\
& GoT            & 2.4 & 2.4 & 2.4 & 2.6 & 3.0 & 3.2 & 3.2 & 3.2 & 2.6 & 2.6 & 2.6 & 2.8 \\
& MechSim        & \textbf{3.8} & \textbf{3.8} & \textbf{3.8} & \textbf{3.4} & \textbf{4.4} & \textbf{4.6} & \textbf{4.6} & \textbf{4.6} & \textbf{4.2} & \textbf{4.2} & \textbf{4.2} & \textbf{4.6} \\
\midrule

\multirow{4}{*}{DeepSeek V3.2}
& Causal & \textbf{4.4} & 3.8 & 3.8 & \textbf{4.6} & 2.4 & 2.2 & 2.2 & 2.4 & 2.0 & 2.0 & 2.0 & 2.2 \\
& Logic       & 1.8 & 2.0 & 2.0 & 1.8 & 1.8 & 3.6 & 2.4 & 2.6 & 3.4 & 3.0 & 3.0 & 3.2 \\
& GoT            & 3.0 & 3.0 & 3.0 & 3.2 & 2.2 & 2.2 & 2.2 & 2.2 & 2.2 & 2.2 & 2.2 & 2.6 \\
& MechSim        & 3.6 & \textbf{4.4} & \textbf{4.2} & 4.2 & \textbf{4.6} & \textbf{4.4} & \textbf{4.4} & \textbf{4.6} & \textbf{4.0} & \textbf{4.0} & \textbf{4.0} & \textbf{4.0} \\
\midrule

\multirow{4}{*}{Qwen 3-235B}
& Causal & 2.8 & 3.0 & 2.8 & 3.0 & 3.4 & 2.2 & 2.4 & 3.6 & 2.6 & 2.2 & 2.2 & 2.6 \\
& Logic       & \textbf{3.6} & 3.0 & 3.0 & \textbf{3.8} & 2.4 & 2.4 & 3.4 & 2.2 & 3.6 & 4.6 & 4.6 & \textbf{4.8} \\
& GoT            & 1.8 & 2.0 & 2.0 & 2.4 & 3.2 & 3.4 & 3.4 & 3.2 & 2.4 & 3.2 & 2.6 & 3.4 \\
& MechSim        & 2.8 & \textbf{3.2} & \textbf{3.2} & 3.2 & \textbf{3.8} & \textbf{3.8} & \textbf{3.8} & \textbf{3.8} & \textbf{4.0} & \textbf{4.7} & \textbf{4.8} & 4.4 \\
\bottomrule
\end{tabular}%
}
\end{table*}

\subsection{Ablation Study Results}
The ablation study (full results in Appendix~\ref{app:ablation}) shows that removing mechanism reasoning causes the largest drop (~50\%), while removing context understanding or knowledge retrieval leads to smaller decreases (~18\%), indicating all components contribute to high-quality explanations, with mechanism reasoning being central, especially for weaker LLMs.

\section{Related Work}

\noindent\textbf{LLMs for Scientific Simulation and Analysis.}
Recent studies increasingly apply LLMs to scientific simulation and analysis~\citep{lu2026towards,baek2025researchagent}.
GenSim exploits LLMs’ grounding and code-generation abilities to construct simulation environments and expert demonstrations~\citep{wang2024gensim}.
G-Sim presents a hybrid approach that integrates LLM-driven structural design with empirical calibration for simulator development~\citep{holt2025gsim}.
However, there is a notable gap in utilizing LLMs as comparative analysts that can synthesize domain knowledge to diagnose logic inconsistencies across multiple scientific simulators~\citep{zecevic2023causal}.

\noindent\textbf{Agentic Scientific Modeling Systems.}
Scientific modeling is increasingly driven by autonomous agentic systems that automate the simulation and experimentation lifecycle. 
For instance, Curie establishes a rigorous framework to manage scientific hypotheses and experimental protocols systematically~\cite{kon2025curie}. 
In public health, agentic systems have been successfully deployed to automate epidemiological model construction~\cite{datta2026agentic} and act as autonomous policymakers to optimize intervention strategies~\cite{aoki2026ai}. 
The AI Scientist introduces an end-to-end pipeline capable of independent idea generation, experimental execution, and automated manuscript preparation~\cite{lu2026towards}.

\section{Conclusion}
This paper introduces MechSim, a neuro-symbolic framework for mechanism-grounded reasoning over executable simulators in simulation-driven decision-making. MechSim combines structured simulator representations, scientific evidence retrieval, sensitivity analysis, and verification-oriented reasoning to generate empirically grounded explanations and recommendations. While MechSim demonstrates strong performance in our benchmarks, extending the framework to increasingly complex scientific domains presents opportunities for future research. Domains such as climate modeling, financial systems, and large-scale infrastructure or biological simulations involve highly interconnected dynamics, substantial uncertainty, and large mechanism spaces that require more scalable reasoning and verification capabilities. Future work could therefore focus on improving MechSim's ability to reason over large-scale dynamical systems, strengthen uncertainty-aware mechanistic inference, and enhance computational efficiency for real-time decision-support applications.

\paragraph{Potential Negative Impacts.}
Although our work aims to improve transparency and grounded reasoning in simulation-driven decision-making, incorrect assumptions, biased observational data, or flawed evidence may still lead to misleading explanations or harmful recommendations. Because the framework may be used in high-stakes domains, overreliance on automatically generated explanations could create risks if users interpret them as guarantees rather than uncertainty-aware decision-support tools. MechSim should therefore be deployed with human oversight, domain-expert validation, and clear communication of uncertainty and system limitations.

\newpage
\bibliographystyle{plain}
\bibliography{references}

\clearpage

\renewcommand{\thesection}{\Alph{section}}
\renewcommand{\thesubsection}{\Alph{section}.\arabic{subsection}}
\renewcommand{\thetable}{\Alph{section}\arabic{table}}
\renewcommand{\thefigure}{\Alph{section}\arabic{figure}}
\title{Simulate, Reason, Decide: Scientific Reasoning with LLMs\\
       for Simulation-Driven Decision Making\\[4pt]
       \large --- Appendix ---}

\author{Anonymous Author(s)}

\maketitle

\appendix

\setcounter{section}{0}
\setcounter{table}{0}
\setcounter{figure}{0}

\tableofcontents
\clearpage

\section{Dataset and Environment Details}
\label{app:dataset}

\setcounter{table}{0}
\setcounter{figure}{0}

\subsection{Datasets}
\label{app:datasets}

We evaluate \mechsim across three domains.
Table~\ref{tab:datasets} summarises the data sources and their key characteristics.

\begin{table}[H]
\centering
\caption{Summary of datasets used across the three experimental domains.}
\label{tab:datasets}
\small
\begin{tabularx}{\linewidth}{p{2.4cm} p{2.8cm} X p{2.2cm}}
\toprule
\textbf{Domain} & \textbf{Dataset} & \textbf{Description} & \textbf{Reference}\\
\midrule
COVID-19
  & JHU CSSE COVID-19 Time Series
  & Daily confirmed cases and deaths aggregated at the US county and national level, covering the period from January 2020 onward.
  & \cite{dong2020interactive}\\[4pt]
Supply Chain
  & M5 Forecasting -- Accuracy
  & Hierarchical daily unit-sales records for 3{,}049 Walmart products across ten stores in three US states (CA, TX, WI), spanning 1{,}941 days (2011--2016). Features include calendar events and sell prices.
  & \cite{makridakis2022m5}\\[4pt]
Measles
  & Measles Scenario Ensemble
  & We conducted our experiment across six configured scenarios (default synthetic, perfect-mixing with and without school-term forcing, London pre-vaccine era, Chicago 2024 migrant shelter, and Texas--New Mexico 2025 outbreak). Population sizes range from 1{,}877 to 2{,}445{,}368; simulation horizons from 35 to 365 days.
  & \cite{masters2024real,allotey2017comparison,gonzalez2025modeling}\\
\bottomrule
\end{tabularx}
\end{table}

\paragraph{COVID-19 data processing.}
we use real-world COVID-19 data from COVID-19 forecast hub, including daily confirmed cases
and deaths aggregated at the US state and national level \cite{cramer2022united}. 
For the forecast task, the pre-cutoff period is used to calibrate each simulator via particle filtering \cite{temfack2025sequential}, and the post-cutoff period serves as the held-out
evaluation window.
We sampled model parameters with random seed and conducted experiments across 50 different locations.

\paragraph{Supply chain data processing.}
The M5 dataset provides hierarchical point-of-sale records at the item--store level for Walmart.
We aggregate daily sales to the product-category and state levels for the supply-chain experiments.

\paragraph{Measles data processing.}
Each scenario is parameterised by a dedicated YAML configuration file specifying population
size~$N$, population structure and immunity/vaccination inputs, contact-matrix structure (flat homogeneous mixing
or a Short Creek--calibrated age-structured matrix), and school-term forcing.
The six scenarios are summarised in Table~\ref{tab:measles_scenarios}.
Oracle traces were loaded as pre-computed aggregate outputs, with one trace for each (config, policy) pair.
Incidence was derived from the daily change in susceptibles  $\Delta S_t$ \cite{grenfell2001travelling}.

\begin{table}[H]
\centering
\caption{Measles scenario configurations used in the scenario ensemble.}
\label{tab:measles_scenarios}
\small
\begin{tabularx}{\linewidth}{p{3.8cm} r r r p{3.6cm}}
\toprule
\textbf{Scenario} & $N$ & $R_0$ & \textbf{Horizon (d)} & \textbf{Notes}\\
\midrule
Default synthetic            &   5{,}630 & 12 & 365 & Short Creek--calibrated age-structured matrix\\
Perfect mixing               &   5{,}630 & 15 & 365 & Flat $1\times1$ matrix; no term forcing\\
Perfect mixing + school term &   5{,}630 & 15 & 365 & Flat matrix; spring/summer/winter closures\\
London pre-vaccine (1944--64) & 2{,}445{,}368 & 28 & 365 & Natural-immunity proxy; UK term forcing\\
Chicago 2024 shelter          &   1{,}877 & 25 & 365 & 5-group demography; flat contact mixing\\
Texas--NM 2025               & 128{,}924 & 30 & 35  & 5-county region represented as one group\\
\bottomrule
\end{tabularx}
\end{table}

% ─── Simulator List ───────────────────────────────────────────────────────────
\subsection{Simulator List}
\label{app:simulators}

\subsubsection{COVID-19 Simulators}
\label{app:covid_sims}

We employ six ODE-based compartmental simulators for the COVID-19 domain, listed in Table~\ref{tab:covid_sims}The simulator dynamics were integrated numerically for a 180-day period to produce trajectories. For the forecast task, the pre-cutoff period is used to calibrate each simulator via particle filtering \cite{temfack2025sequential}, and the post-cutoff period serves as the held-out

\begin{table}[H]
\centering
\caption{COVID-19 simulators used in the evaluate and forecast tasks.}
\label{tab:covid_sims}
\small
\begin{tabularx}{\linewidth}{p{2.2cm} p{3.0cm} X p{2.5cm}}
\toprule
\textbf{Model} & \textbf{Compartments} & \textbf{Key Mechanisms \& Assumptions} & \textbf{Reference}\\
\midrule
SIR
  & S, I, R
  & Homogeneous mixing; constant $\beta$ and $\gamma$; permanent immunity; no latency, mortality, or hospitalisation.
  & \cite{kermack1927contribution,hethcote2000mathematics}\\[4pt]

SEIRD
  & S, E, I, R, D
  & Exposed latency ($\sigma = 1/\text{incubation}$) and disease-induced mortality ($\mu$); homogeneous mixing.
  & \cite{hethcote2000mathematics}\\[4pt]

SEIR-HD
  & S, E, I, R, H, D
  & Explicit hospitalisation compartment ($H$) with capacity threshold; separates community mortality ($\mu_I$) from hospital CFR ($\delta$); enables overload tracking.
  & \cite{ferguson2020report,moghadas2020projecting}\\[4pt]

SEIQRD
  & S, E, I, Q, R, D
  & Explicit quarantine compartment ($Q$); quarantine rate $q$ directly reduces effective transmission.
  & \cite{tang2020estimation}\\[4pt]

SEAIHRD
  & S, E, A, I, H, R, D
  & Asymptomatic transmission ($A$) with relative transmissibility $\rho \approx 0.5$; fraction $f \approx 0.3$ of exposed become asymptomatic.
  & \cite{he2020temporal}\\[4pt]

SIRV
  & S, I, R, V, D
  & Vaccination compartment ($V$) with daily coverage rate $\nu$; vaccinated individuals fully protected.
  & \cite{diekmann2013mathematical}\\
\bottomrule
\end{tabularx}
\end{table}

\subsubsection{Supply Chain Simulators}
\label{app:sc_sims}

Four supply chain simulators are used in the supply chain domain (Table~\ref{tab:sc_sims}). All operate in discrete time with Poisson external demand. Costs include per-unit holding cost $c_h$ and per-unit backlog cost $c_b$.

\begin{table}[H]
\centering
\caption{Supply chain simulators used in the evaluate and forecast tasks.}
\label{tab:sc_sims}
\small
\begin{tabularx}{\linewidth}{p{2.5cm} X p{2.8cm}}
\toprule
\textbf{Model} & \textbf{Key Mechanisms \& Assumptions} & \textbf{Reference}\\
\midrule
Beer Game (BG)
  & Four-echelon chain (retailer $\to$ wholesaler $\to$ distributor $\to$ factory) with Sterman anchor-and-adjust heuristic ordering. Information and shipping delays of 2 periods each; captures the bullwhip effect.
  & \cite{sterman1989modeling}\\[4pt]

Base Stock (BS)
  & Single-stage, periodic review. Orders up to a fixed base-stock level $S$ each period; deterministic lead time $L$. Optimal under linear holding and backlog costs.
  & \cite{porteus2002foundations}\\[4pt]

$(s,S)$ Inventory
  & Orders $S - I_t$ only when inventory falls below reorder point $s$; fixed ordering cost $K$. Captures economies of scale in ordering.
  & \cite{scarf1960optimality}\\[4pt]

Multi-Echelon (ME)
  & Decentralised three-level chain with independent $(r,Q)$ policies at each level; deterministic lead times. Captures bullwhip amplification across echelons.
  & \cite{porteus2002foundations}\\
\bottomrule
\end{tabularx}
\end{table}

\subsubsection{Measles Simulators}
\label{app:measles_sims}

Five compartmental baseline models are used as candidate simulators for the measles environment (Table~\ref{tab:measles_sims}), implemented via the epydemix platform \cite{gozzi2025epydemix} and calibrated with Approximate Bayesian Computation (ABC)~\citep{beaumont2002approximate}.

\begin{table}[H]
\centering
\caption{Measles simulators used in the evaluate and forecast tasks.}
\label{tab:measles_sims}
\small
\begin{tabularx}{\linewidth}{p{2.2cm} p{3.2cm} X p{2.4cm}}
\toprule
\textbf{Model} & \textbf{Compartments} & \textbf{Key Mechanisms \& Assumptions} & \textbf{Reference}\\
\midrule
SIR
  & S, I, R
  & Permanent immunity post-infection; $R_0 \approx 12$--$18$ for measles.
  & \cite{kermack1927contribution,hethcote2000mathematics}\\[4pt]

SEIR
  & S, E, I, R
  & Explicit latent period ($\sigma$); captures the $\sim$8--12 day measles incubation.
  & \cite{diekmann2013mathematical}\\[4pt]

SIRV
  & S, I, R, V
  & Single-dose vaccination ($\nu$); susceptibles move to protected class.
  & \cite{diekmann2013mathematical}\\[4pt]

SVEIR
  & S, V$_1$, V$_2$, E, I, R
  & Two-dose vaccination schedule with waning immunity; partial protection after dose 1, full after dose 2.
  & \cite{diekmann2013mathematical}\\[4pt]

SVVIR
  & S, V$_1$, V$_2$, I, R
  & Compressed two-dose structure without explicit exposed state; used when latent-period data are unavailable.
  & \cite{bauch2004vaccination}\\
\bottomrule
\end{tabularx}
\end{table}

% ─── Oracle Models ────────────────────────────────────────────────────────────
\subsection{Oracle Models}
\label{app:oracle}

\subsubsection{COVID-19 Oracle}
\label{app:covid_oracle}

The COVID-19 Oracle is a stochastic, multi-compartment discrete-time simulator \cite{kain2021chopping} designed to capture epidemic dynamics beyond the representational capacity of the simple ODE models evaluated by MechSim. It serves as the ground-truth reference for the policy selection task.

\paragraph{Compartment structure.}
$S$ (susceptible), $E$ (latent), $I_a$ (asymptomatic), $I_p$ (pre-symptomatic), $I_m$ (mild symptomatic), $I_s$ (severe symptomatic), $H_r$ (hospitalised, recovering), $H_d$ (hospitalised, dying), $R$ (recovered), $D$ (dead). Daily incident counters $D_\text{new}$, $H_\text{new}$, $I_\text{new,sympt}$ reset each period.

\paragraph{Transition dynamics.}
All transitions are stochastic, drawn from Binomial/Multinomial distributions using an Euler-Multinomial scheme with sub-day step $\Delta t = 1/6$:
\begin{align}
  \Delta S \to E &\sim \mathrm{Binomial}\!\left(S,\, 1 - e^{-\beta_t \Lambda \Delta t}\right), \quad
  \Lambda = \frac{C_a I_a + C_p I_p + \mathrm{iso}_m C_m I_m + \mathrm{iso}_s C_s I_s}{N}.
\end{align}

\paragraph{Key parameters.}
Fraction asymptomatic $\alpha = 0.30$; relative contact weights $C_a{=}0.50$, $C_p{=}0.80$, $C_m{=}0.60$, $C_s{=}1.00$; latent period $T_\text{lat} = 4$ days; probability of mild given symptomatic $\mu = 0.50$; hospitalisation fraction from severe $\delta = 0.12$; hospital CFR $\rho_d^{-1} = 9$ days; hospital recovery $\rho_r^{-1} = 14$ days.

\paragraph{Harm function.}
\begin{equation}
  \mathrm{Harm} = C_\text{death} \cdot D_\text{total} + C_\text{overload} \cdot H_\text{overload} + \lambda_\text{policy} \cdot c_\text{policy} \cdot T,
\end{equation}
where $C_\text{death} = 10{,}000$, $C_\text{overload} = 100$ (per overload day), $\lambda_\text{policy} = 5$. Harm is normalised per 100{,}000 population. Monte Carlo: $M = 500$ replicates per (scenario, intervention) pair.

\subsubsection{Supply Chain Oracle}
\label{app:supply_chain_oracle}

The Supply Chain Oracle is an agent-based, discrete-time stochastic simulator \cite{chu2015simulation} designed to capture the complex, dynamic behaviors of a distributed inventory network. It evaluates the performance of decentralised $(r, Q)$ inventory control policies across interacting facilities under demand and lead-time uncertainties.

\paragraph{Network structure.}
The system models a divergent three-echelon distribution network. It consists of one plant (acting as an infinite-inventory root), one central warehouse, two distribution centers, and four customer-facing distributors. Facilities interact via forward material flows (shipments) and backward information flows (orders).

\paragraph{Transition dynamics.}
The simulator operates on a discrete daily time step ($t = 1, \dots, N_T$). Each facility agent autonomously monitors its inventory position, defined as on-hand inventory minus backorders plus on-order inventory. The transition logic dictates that when a facility's inventory position falls to or below its reorder point $r_i$, it places a replenishment order of quantity $Q_i$ ($ROI_{it} = 1$). 
Unmet demands are strictly backordered. Order processing times follow truncated uniform distributions, but the actual cumulative lead time dynamically depends on upstream inventory availability: if the upstream facility is stocked out, it must first replenish its own inventory before processing the downstream order.

\paragraph{Key parameters.}
Daily customer order demands follow Gaussian distributions (with means ranging from 400 to 1000, and standard deviations from 200 to 500). Order processing times across edges are bounded uniform distributions (e.g., $[10,15]$, $[8,12]$, $[4,6]$ days). Facility-level cost heterogeneity is modeled with a uniform holding cost of $1$ m.u./(unit$\cdot$day) , and varying fixed reorder costs: $300$ m.u. for the warehouse, $200$ m.u. for distribution centers, and $100$ m.u. for distributors.

\paragraph{Objective function (Harm).}
While the original framework enforces a $\ge 95\%$ fill rate as a probability constraint, the Oracle adapts this into a penalised composite scalar, \textit{Harm}:
$$
\mathrm{Harm} = \mathrm{INVC} + C_\text{stockout} \cdot \mathrm{UnmetDemand} + \lambda_\text{policy} \cdot c_\text{policy} \cdot T,
$$
where the base inventory cost ($\mathrm{INVC}$) over the horizon $N_T$ is calculated as the sum of average on-hand holding costs and discrete reorder setup costs:
$$
\mathrm{INVC} = \sum_{i=1}^{N_F} c_i^H \mathrm{AIO}_i + \sum_{i=1}^{N_F} c_i^R \mathrm{SRO}_i.
$$
The $C_\text{stockout}$ penalty replaces the strict service-level constraints for continuous policy evaluation. The expectations of these stochastic performance functions are estimated via the Monte Carlo method with $M = 300$ replicates per intervention.

\subsubsection{Measles Oracle}
\label{app:measles_oracle}

The measles Oracle is a stochastic agent-based \texttt{measles::ModelMeaslesMixing} simulator accessed through \texttt{rpy2}/epiworldR~\cite{epiworldR2023,R-measles}, matching the oracle definition used in the main benchmark description. For policy evaluation, we use precomputed aggregate traces from this R-backed Oracle for each configuration--policy pair; horizons follow the scenario YAMLs, with five configurations running days 0--365 and Texas--NM 2025 running days 0--35. The incidence target is stored in the \texttt{target} column and is computed from daily depletion of \texttt{Susceptible} plus \texttt{Quarantined\_Susceptible}, clipped at zero.

% ─── Policy Lists ─────────────────────────────────────────────────────────────
\subsection{Intervention Policy Lists}
\label{app:policies}

\subsubsection{COVID-19 Policies}
The table summarizes the intervention policies used in the policy selection task, with each parameter defined following \cite{brauner2021inferring,moore2021vaccination}.

\begin{table}[H]
\centering
\caption{COVID-19 intervention policies for the policy selection task.}
\label{tab:covid_policies}
\small
\begin{tabularx}{\linewidth}{p{0.55cm} p{3.3cm} p{1.1cm} p{1.2cm} X}
\toprule
\textbf{ID} & \textbf{Title} & \textbf{Cost} & \textbf{$R_0$} & \textbf{Mechanism}\\
\midrule
A1 & No Intervention   & 0.0  & 2.6  & Baseline; no modifications.\\
A2 & Mass Testing Only & 0.10 & 2.4  & Detection without isolation; mild WFH ($\times0.95$ contact).\\
A3 & Strict Lockdown   & 0.80 & 0.7  & Full contact reduction ($\times0.25$) with case isolation.\\
A4 & Elderly Vaccination & 0.30 & 1.8 & Targeted high-risk vaccination; reduces mortality not transmission.\\
A5 & Masking + WFH     & 0.20 & 1.6  & Moderate contact reduction ($\times0.65$); no isolation.\\
A6 & Testing + Isolation & 0.50 & 1.1 & SIP ($\times0.55$) + isolation of mild ($\times0.15$) and severe ($\times0.08$) cases.\\
\midrule
\multicolumn{5}{l}{\textit{Stress-test policies:}}\\
\midrule
A7 & Reduce Trans.\ $+$ Hospital Overload & 0.35 & 1.2 & Transmission $\downarrow$ but Oracle raises hospitalisation fraction ($\delta$: $0.12{\to}0.20$).\\
A8 & Delay Peak $+$ Increase Deaths & 0.25 & 1.5 & Slows recovery; Oracle accelerates hospital death rate.\\
A9 & Mild Cases Only   & 0.15 & 2.2  & Isolates mild cases only; severe transmission unchanged in Oracle.\\
\bottomrule
\end{tabularx}
\end{table}

\subsubsection{Supply Chain Policies}

\begin{table}[H]
\centering
\caption{Supply chain intervention policies for the policy selection task.}
\label{tab:sc_policies}
\small
\begin{tabularx}{\linewidth}{l l l X}
\toprule
\textbf{ID} & \textbf{Title} & \textbf{Cost} & \textbf{Mechanism} \\
\midrule
P1 & No Intervention (Baseline)       & 0.0  & Maintain calibrated $(r,Q)$ baseline. \\
P2 & Safety-Stock Boost (+50\%)      & 0.25 & Reorder point $r$ raised by 50\% at all facilities. \\
P3 & Lead-Time Reduction (--30\%)    & 0.40 & Processing times on all edges cut by 30\%. \\
P4 & Demand Sharing / VMI             & 0.20 & Demand std reduced 20\% at distributors via information sharing. \\
P5 & Order Quantity +30\% (EOQ)      & 0.15 & Order quantity $Q$ raised 30\%; reduces ordering frequency. \\
P6 & Expedited Replenishment          & 0.55 & Emergency reorder at $2\times r$; near-eliminates stockouts. \\
\bottomrule
\end{tabularx}
\end{table}

\subsubsection{Measles Policies}

\begin{table}[H]
\centering
\caption{Measles intervention policies for the policy selection task.}
\label{tab:measles_policies}
\small
\begin{tabularx}{\linewidth}{p{0.55cm} p{3.2cm} p{1.1cm} X}
\toprule
\textbf{ID} & \textbf{Policy} & \textbf{Cost} & \textbf{Mechanism}\\
\midrule
A1 & No Intervention & 0.0 & Baseline with quarantine, isolation, and contact tracing disabled.\\
A2 & Isolation Only & 0.10 & Symptomatic individuals self-isolate; quarantine and contact tracing are disabled.\\
A3 & Quarantine + Contact Tracing & 0.45 & Contact tracing identifies exposures and quarantines traced contacts; self-isolation is disabled.\\
A4 & School Closure & 0.35 & Scales the contact matrix to 50\% of baseline contacts without quarantine, isolation, or tracing.\\
A5 & Full Response & 0.85 & Combines isolation, quarantine, contact tracing, and school-closure contact reduction.\\
A6 & Enhanced Full Response & 1.10 & Full response with no detection delay and a longer symptomatic isolation period.\\
A7 & Vaccination Campaign & 0.60 & Increases configured per-group vaccination rates by a 1.10 multiplier, capped at 1.0.\\
\bottomrule
\end{tabularx}
\end{table}

% ═══════════════════════════════════════════════════════════════════════════════
\section{Experiment Details}
\label{app:experiments}
% ═══════════════════════════════════════════════════════════════════════════════

\setcounter{table}{0}
\setcounter{figure}{0}

\subsection{Baseline Method Details}
\label{app:baselines}

All baselines receive identical simulator outputs and produce structured explanations in the $\mathcal{E} = (\mathcal{I}, \mathcal{P}, \mathcal{Z}, \mathcal{C})$ schema.

\paragraph{Causal-Copilot~\citep{wang2025causal}.}
An autonomous causal analysis agent that operationalises expert-level causal discovery within an LLM framework. We faithfully reimplement its four-phase pipeline:
\textit{(i)} Algorithm selection — an LLM characterises the data regime and selects from 20+ causal discovery algorithms (PC, FCI, GES, LiNGAM, NOTEARS, etc.);
\textit{(ii)} Causal graph construction — the selected algorithm is run on simulation output data;
\textit{(iii)} Causal effect estimation via do-calculus and regression adjustment;
\textit{(iv)} LLM interpretation grounded in the learned graph.
This differs from \mechsim in that Causal-Copilot discovers causal structure from observational data, whereas \mechsim formalises the known mechanistic structure of each simulator via Mechanism Graphs.

\paragraph{Logic-LM~\citep{pan2023logic}.}
Integrates symbolic reasoning with LLMs for faithful logical inference via three stages:
\textit{(i)} Symbolic formalisation — the LLM translates simulation summaries into first-order-logic predicates and transition rules;
\textit{(ii)} Constraint checking — a lightweight Python solver evaluates the rule set, derives a canonical simulator ranking, and flags logical contradictions;
\textit{(iii)} LLM interpretation grounded in verified logical derivations.
Logic-LM provides structured conclusions but lacks grounding in domain-specific mechanistic assumptions.

\paragraph{Graph of Thoughts (GoT)~\citep{besta2024graph}.}
Structures inference as a directed graph of thought nodes. We implement three GoT operations: \textit{Generate} — spawn independent thought nodes per simulator and per outcome metric; \textit{Aggregate} — merge compatible thoughts to identify common mechanistic drivers; \textit{Refine} — improve the aggregated thought using retrieved scientific evidence. Unlike \mechsim, GoT does not constrain reasoning paths to verified mechanistic graph transitions.

\subsection{Calibration Methods}
\label{app:calibration}

\paragraph{COVID-19 Policy Selection Task.}
Simulators are parameterised using the default scenario schema (Table~\ref{tab:covid_default_params}); parameters are set from published epidemiological estimates. We sampled the parameters across 100 scenarios to test the robustness of our method.

\begin{table}[H]
\centering
\caption{Default COVID-19 scenario parameter ranges for the evaluation and policy selection task. Values indicate the ranges explored in simulations.}
\label{tab:covid_default_params}
\small
\begin{tabularx}{\linewidth}{l X}
\toprule
\textbf{Parameter} & \textbf{Value Range} \\
\midrule
Population $N$ & log-uniform between $100{,}000$ and $2{,}000{,}000$ \\
Basic reproduction number $R_0$ & uniform between $2.0$ and $3.5$ \\
Infection duration (days) & uniform between $1.0$ and $6.0$ \\
Latent period (days) & uniform between $3.0$ and $12.0$ \\
Initial infected $I_0$ & log-uniform between $50$ and $2{,}000$ \\
Hospital bed rate (per 1k population) & uniform between $1.5$ and $10.0$ \\
Simulation horizon (days) & choice among $90$, $120$, $180$, or $365$ \\
\bottomrule
\end{tabularx}

\begin{flushleft}
\footnotesize
Parameters were set according to publicly available data and guidelines: 
\cite{who_covid_dashboard} and \cite{cdc_covid_yellowbook}.
\end{flushleft}
\end{table}

\paragraph{COVID-19 forecast task.}
Each ODE simulator is calibrated to observed US weekly death data using particle filtering \cite{temfack2025sequential}. The objective minimises MAE normalised by
mean observed cases on the training split. Settings: strategy \texttt{best1bin};
population $15\times(\text{num.\ params})$; max iterations 200; tolerance $10^{-7}$;
seed 42; polish enabled. Candidate simulators are SIR, SEIRD, SEIRHD, SEIQRD,
SEAIHRD, and SIRV, each with their respective free parameters and logistic
time-varying transmission rate $\beta(t)$.

\paragraph{Supply chain policy task.}
Two-step calibration: \textit{(i)} run the Oracle for 28 periods with 5 Monte Carlo
replicates under no-intervention (P1) to obtain ``historical'' demand and stockout
series from the M5 dataset~\citep{makridakis2022m5}; \textit{(ii)} fit each
simulator's parameters to match observed demand mean, standard deviation, and
lead-time distribution. Candidate simulators include base-stock inventory control,
$(s,S)$ inventory control, decentralised multi-echelon $(r,Q)$ control, and the
Beer Game supply-chain model. Policy effects are modelled as covariate overrides
(reorder-point multipliers, lead-time multipliers, demand multipliers) applied on
top of the calibrated baseline.

\paragraph{Measles forecast task.}
Simulators are calibrated via Approximate Bayesian Computation
(ABC)~\citep{beaumont2002approximate} using \texttt{epydemix}'s
\texttt{ABCSampler}. Candidate simulators are SIR, SEIR, SIRV, SVEIR, and SVVIR,
implemented with Epydemix~\citep{kuddus2021mathematical}. The first 14 days of
incidence are used for fitting; the remaining held-out horizon is then forecast.
Prior distributions (all uniform): initial infected, bounded by early incidence;
initial exposed for SEIR and SVEIR; transmission rate $\mathcal{U}(0.05, 1.50)$;
recovery rate $\mathcal{U}(0.05, 0.50)$; incubation rate (SEIR, SVEIR)
$\mathcal{U}(0.03, 0.33)$; vaccination rate, second-dose rate, and protection rate
(SIRV, SVEIR, SVVIR) $\mathcal{U}(0.0, 0.08)$; waning rate
$\mathcal{U}(0.0, 0.02)$.

\subsection{Parameter Settings}
\label{app:hyperparams}
\begin{table}[H]
\centering
\caption{Key hyperparameter settings across all tasks and domains.}
\label{tab:hyperparams}
\small
\begin{tabularx}{\linewidth}{l l X}
\toprule
\textbf{Component} & \textbf{Parameter} & \textbf{Value} \\
\midrule
\multirow{3}{*}{LLM} 
 & Main model & claude-sonnet-4.6, DeepSeek V3.2, Qwen3-235B \\
 & Max tokens (explanations) & 2{,}000 \\
 & Temperature (reasoning / judge) & 0.7 / 0.0 \\
\midrule
\multirow{3}{*}{Knowledge Retrieval} 
 & API & arXiv, PubMed \\
 & Re-ranker & Cross-encoder (two-stage, inspired by OpenScholar~\cite{asai2026synthesizing}) \\
 & Top-$K$ papers & 5; maximum abstract 800 characters \\
\midrule
\multirow{2}{*}{Simulation} 
 & Number of environments & 3 \\
 & Samples per environment & 100 per intervention \\
\midrule
\multirow{3}{*}{Calibration} 
 & Max iterations & 200 \\
 & Population size & $15 \times$ (number of parameters) \\
 & Tolerance & $10^{-7}$ \\
\midrule
\multirow{4}{*}{Evaluation} 
 & Judge LLM rounds & 5 independent rounds \\
 & Explanation dimensions & 4 (completeness, mechanistic logic consistency, scientific soundness, faithfulness) \\
 & Scoring scale & 1--5 per dimension \\
 & Policy metrics & Precision@3, Precision@5, Recall@3, Recall@5 \\
\midrule
\multirow{2}{*}{Sensitivity Analysis} 
 & Parametric variants & 4--5 values per parameter \\
 & Structural variant types & Mixing pattern, immunity model, severity structure \\
\bottomrule
\end{tabularx}
\end{table}

\subsection{Prompt Templates}
\label{app:prompts}

We present the key prompt templates used by each \mechsim agent. Placeholders \texttt{\{...\}} are populated at runtime.

\subsubsection{Context Understanding Agent}

\begin{tcolorbox}[
  title={Prompt: Context Understanding Agent (Epidemic Domain)},
  colback=promptbg, colframe=promptframe,
  fonttitle=\small\bfseries, fontupper=\small,
  breakable, enhanced
]
\small
\textbf{System:} You are an expert in epidemiological modeling and public health policy. Analyze the following epidemic scenario to construct a structured semantic context.

\vspace{4pt}
\textbf{Task:} Perform three functions simultaneously:
\begin{enumerate}[noitemsep,leftmargin=*]
\item \textbf{Environment Definition}: Identify real-world factors (population traits, healthcare capacity, geographic context) that constrain model assumptions and mechanisms.
\item \textbf{Goal Identification}: Specify the decision-making objective (policy evaluation or forecasting).
\item \textbf{Key Entity Recognition}: Extract critical variables from the scenario ($R_0$, $\beta$, $\gamma$, hospital beds, population).
\end{enumerate}

\textbf{[Scenario Specification]}\\
Population: \texttt{\{N\}};\ Initial Infected: \texttt{\{I0\}};\ R0: \texttt{\{R0\}};\ Hospital Beds: \texttt{\{hospital\_beds\}};\ Horizon: \texttt{\{horizon\}} days;\ Task: \texttt{\{task\}}

\vspace{4pt}
Return ONLY valid JSON with keys: \texttt{environment} (geographic\_context, healthcare\_capacity, real\_world\_factors), \texttt{goal} (primary\_objective, decision\_horizon, key\_outcomes), \texttt{key\_entities}, \texttt{assumption\_validity\_notes}.
\end{tcolorbox}

\subsubsection{Mechanism Modeling Agent}
\begin{tcolorbox}[
  title={Prompt: Mechanism Graph Extraction},
  colback=promptbg, colframe=promptframe,
  fonttitle=\small\bfseries, fontupper=\small,
  breakable, enhanced
]
\small
You are an expert in scientific simulation, compartmental modeling, and causal
inference. Your task is to extract a complete, structured mechanism graph
$\mathcal{G}_i = (\mathcal{V}_i, \mathcal{E}_i, \mathcal{M}_i)$ for the
\texttt{\{model\_name\}} simulator, where nodes represent state variables or
simulator entities, edges represent mechanistic transitions, and metadata captures
assumptions, execution traces, and summary statistics.

\vspace{4pt}
\textbf{[Simulator Definition]}\ \texttt{\{simulator\_source\_or\_description\}}

\textbf{[Execution Traces]}\ \texttt{\{execution\_traces\}}

\textbf{[Simulation Summary Statistics]}
Peak infection rate: \texttt{\{peak\_infection\_rate\}};
Total deaths: \texttt{\{total\_deaths\}};
Hospital overload days: \texttt{\{hospital\_overload\_days\}};
Additional outputs: \texttt{\{extra\_summary\_stats\}}

\textbf{[Scientific Evidence]}\ \texttt{\{retrieved\_evidence\}}

\vspace{4pt}
Extract the mechanism graph according to the following schema:
\begin{enumerate}[leftmargin=*, itemsep=2pt]
  \item \textbf{State nodes ($\mathcal{V}_i$):} List all simulator compartments or
  state variables (e.g., S, E, I, R, H, D, V). Each node must be a plain string
  matching the simulator's variable names exactly.
  \item \textbf{Mechanistic edges ($\mathcal{E}_i$):} For each transition, specify:
  \texttt{from}, \texttt{to} (plain strings); \texttt{mechanism} (the rate or
  process driving the transition, e.g., $\beta \cdot S \cdot I / N$);
  \texttt{activated\_by} (the simulator assumption in $\mathcal{A}_i$ that enables
  this transition, e.g., homogeneous mixing, waning immunity).
  \item \textbf{Graph metadata ($\mathcal{M}_i$):} Extract the following:
  \begin{itemize}[itemsep=1pt]
    \item \texttt{assumptions} $\mathcal{A}_i$: list all structural assumptions
    encoded by this simulator (e.g., permanent immunity, constant $\beta$,
    homogeneous mixing);
    \item \texttt{execution\_summary}: key statistics attached as graph-level
    metadata (peak infections, mortality, overload days);
    \item \texttt{reasoning\_path}: a topologically ordered sequence of nodes
    representing the primary propagation pathway from initial state to
    decision-relevant outcome;
    \item \texttt{dot\_code}: a Graphviz-compatible DAG representation of the
    full graph;
    \item \texttt{scientific\_summary}: a brief statement grounding the simulator's
    assumptions and mechanisms in retrieved scientific evidence.
  \end{itemize}
\end{enumerate}

\vspace{4pt}
\textbf{Important constraints:}
\begin{itemize}[leftmargin=*, itemsep=1pt]
  \item Every \texttt{from}/\texttt{to} field must be a plain string; no nested
  objects or numeric indices.
  \item Do not introduce nodes or edges absent from the simulator definition or
  execution traces.
  \item If an assumption is absent or cannot be verified from the simulator source,
  explicitly mark it as \texttt{unverified} in $\mathcal{A}_i$.
\end{itemize}

\vspace{4pt}
Return ONLY valid JSON with the following top-level keys:
\texttt{assumptions}, \texttt{nodes}, \texttt{edges}, \texttt{execution\_summary},
\texttt{reasoning\_path}, \texttt{dot\_code}, \texttt{scientific\_summary}.
\end{tcolorbox}

\subsubsection{Mechanism Reasoning Agent}
\begin{tcolorbox}[
  title={Prompt: Mechanism Reasoning Agent},
  colback=promptbg, colframe=promptframe,
  fonttitle=\small\bfseries, fontupper=\small,
  breakable, enhanced
]
\small
\textbf{Task:} You are an expert in scientific simulation and mechanism-grounded
reasoning. Your goal is to produce a structured, verifiable explanation of the
simulation results by reasoning strictly over the extracted mechanism graphs,
execution traces, and retrieved scientific evidence. Do \textbf{not} summarize
outputs superficially; instead, trace how simulator assumptions activate
mechanistic propagation paths that produce the observed outcomes.

\vspace{4pt}
\textbf{[Simulation Results]}\ \texttt{\{sim\_summaries["models"]\}}

\textbf{[Scenario Context]}\ Environment: \texttt{\{geographic\_context\}};
Objective: \texttt{\{primary\_objective\}};
Assumption validity notes: \texttt{\{assumption\_validity\_notes[:2]\}}

\textbf{[Mechanism Graphs]}\ \texttt{\{mechanism\_graphs\}}

\textbf{[Execution Traces]}\ \texttt{\{execution\_traces\}}

\textbf{[Scientific Evidence (RAG)]}\ \texttt{\{retrieved\_evidence\}}

\textbf{[Sensitivity Analysis Findings]}\ Top drivers:
\texttt{\{top\_3\_drivers\}}; Validated hypotheses:
\texttt{\{validated\_hypotheses[:2]\}}.
Use these findings to confirm, refine, or qualify each mechanistic claim.

\vspace{4pt}
\textbf{[Mechanism Reasoning Constraints]}
\begin{itemize}[leftmargin=*, itemsep=1pt]
  \item You MUST follow the mechanism graph strictly. Each reasoning step must
  correspond to a valid node-to-node transition in the graph.
  \item Do NOT skip intermediate states or introduce variables absent from the graph.
  \item When evidence is incomplete, conflicting, or insufficient to support a claim,
  explicitly represent the uncertainty rather than asserting unsupported conclusions.
  \item Cross-reference sensitivity analysis findings at each step: if a transition
  corresponds to a top-ranked driver, explicitly note its empirical effect size.
\end{itemize}

\vspace{2pt}
For each simulator, trace the mechanism path step by step, e.g.:
\begin{verbatim}
[Model: SEIR-HD — Mechanism Path]
  Step 1 (S): Explain initial susceptible pool and inflow conditions.
  Step 2 (E): Describe S→E transition; identify the activating assumption.
  Step 3 (I): Describe E→I transition; link to sensitivity-ranked drivers.
  ...
  Step N (H/D): Explain terminal state; assess real-world implications.
\end{verbatim}

\vspace{4pt}
\textbf{[Structured Explanation --- $\mathcal{E} = (\mathcal{I}, \mathcal{P},
\mathcal{Z}, \mathcal{C})$]}

\textbf{1.~Output Interpretation ($\mathcal{I}$):} Synthesize the scenario context,
task objective, and simulator outputs. Identify decision-relevant patterns (e.g.,
peak divergence, mortality gaps, capacity breaches) and connect them to real-world
implications for the deployment context.

\textbf{2.~Mechanism Reasoning Paths ($\mathcal{P}$):} For each simulator, trace
the full propagation path node-by-node. For each transition, explicitly state: (a)
the mechanism label on the edge, (b) the assumption in $\mathcal{A}_i$ that activates
it, and (c) whether sensitivity analysis confirms it as a key driver.

\textbf{3.~Supporting Evidence ($\mathcal{Z}$):} For each claim, cite retrieved
scientific evidence with specific quantitative findings. Where evidence conflicts
with simulator predictions, explicitly flag the discrepancy and assess its impact
on reliability.

\textbf{4.~Claims ($\mathcal{C}$):} State 3--5 mechanism-grounded claims. Each
claim must: (a) identify the responsible simulator assumption, (b) trace the full
propagation path through $\mathcal{P}$, (c) cite a specific evidence reference from
$\mathcal{Z}$, and (d) note any uncertainty or assumption-context mismatch that
limits confidence.

\textbf{5.~Decision Recommendation ($\mathcal{R}$):} Provide actionable,
mechanism-grounded recommendations for the decision maker.  All
recommendations must be consistent with the verified explanation and finalized only
after the full reasoning chain is complete.
\end{tcolorbox}

\subsubsection{Policy Selection Prompt}
\begin{tcolorbox}[
  title={Prompt: Policy Selection},
  colback=promptbg, colframe=promptframe,
  fonttitle=\small\bfseries, fontupper=\small,
  breakable, enhanced
]
\small
You are an expert scientific advisor specializing in simulation-driven decision-making.
Your task is to recommend the \textbf{top-3 intervention policies} from the candidate
list, ranked from most to least effective, based on simulator outputs, mechanism graphs,
sensitivity analysis findings, and retrieved scientific evidence.

\vspace{4pt}
\textbf{[Context Information]}\ \texttt{\{context\_summary\}}

\textbf{[Candidate Policies]}\ \texttt{\{policy\_list\_with\_descriptions\}}

\textbf{[Calibrated Simulator Predictions]}\ \texttt{\{simulation\_result\}}

\textbf{[Mechanism Graphs]}\ \texttt{\{mechanism\_graphs\}}

\textbf{[Sensitivity Top Drivers]}\ \texttt{\{sensitivity\_ranking[:3]\}}

\textbf{[Scientific Evidence]}\ \texttt{\{retrieved\_evidence\}}

\vspace{4pt}
Reason step by step for each candidate policy:
\begin{enumerate}[leftmargin=*, itemsep=1pt]
  \item \textbf{Mechanism tracing}: Identify which nodes and edges in the mechanism
  graph this policy activates or suppresses. Trace the full propagation path from
  intervention to outcome (e.g., policy $\rightarrow$ contact rate $\beta\downarrow$
  $\rightarrow$ force of infection $\downarrow$ $\rightarrow$ peak infections
  $\downarrow$).
  \item \textbf{Simulator faithfulness}: Assess whether each candidate simulator
  structurally captures the mechanisms this policy targets. Flag simulators that
  lack the relevant compartments or assumptions needed to model the policy's effects.
  \item \textbf{Sensitivity alignment}: Cross-reference the top sensitivity drivers
  with the policy's mechanism. Prioritize policies that act on the highest-ranked
  drivers.
  \item \textbf{Evidence grounding}: Cite retrieved scientific evidence that supports
  or challenges the expected effectiveness of each policy under the given deployment
  context.
  \item \textbf{Assumption validity}: Explicitly note whether any simulator assumptions
  (e.g., homogeneous mixing, permanent immunity) may limit the reliability of
  predictions for this policy.
\end{enumerate}

\vspace{4pt}
Return ONLY valid JSON with the following fields:\\
\texttt{ranking}: ordered list of top-3 policy IDs (best to worst);\\
\texttt{reasoning}: per-policy justification grounded in mechanism paths and evidence;\\
\texttt{mechanism\_paths}: key propagation paths activated by each ranked policy;\\
\texttt{assumption\_flags}: simulator assumptions that may affect prediction reliability;\\
\texttt{confidence}: per-policy confidence score (low / medium / high) with rationale.
\end{tcolorbox}

\subsubsection{Verification Agent Prompt}
\begin{tcolorbox}[
  title={Prompt: Verification Agent},
  colback=promptbg, colframe=promptframe,
  fonttitle=\small\bfseries, fontupper=\small,
  breakable, enhanced
]
\small
You are a scientific validator for simulation-driven explanations. Given the
mechanism graph $\mathcal{G}_i$, execution traces, and generated explanation
$\mathcal{E} = (\mathcal{I}, \mathcal{P}, \mathcal{Z}, \mathcal{C})$ below,
perform three verification checks.

\vspace{4pt}
\textbf{[Mechanism Graph]}\ \texttt{\{mechanism\_graph\}} \quad
\textbf{[Execution Traces]}\ \texttt{\{execution\_traces\}} \quad
\textbf{[Generated Explanation]}\ \texttt{\{explanation\}} \quad
\textbf{[Sensitivity Results]}\ \texttt{\{sensitivity\_results\}}

\vspace{4pt}
\textbf{1. Structural Consistency:} (a) Do all nodes match valid compartments for
\texttt{\{model\_name\}}? (b) Are edges mechanistically valid with no impossible
transitions (e.g., D$\to$S, S$\to$R without infection)? (c) Is each reasoning path
in $\mathcal{P}$ a valid topological ordering of $\mathcal{G}_i$?

\textbf{2. Evidence Grounding:} (a) Are all $\mathcal{E}$-components present and
substantive? (b) Is each claim in $\mathcal{C}$ traceable to a valid path in
$\mathcal{P}$ and grounded in $\mathcal{Z}$, with no hallucinated variables or
unsupported effect directions? (c) Are empirical findings from sensitivity analysis
reflected where relevant?

\textbf{3. Empirical Support:} Are recommendations in $\mathcal{R}$ actionable,
consistent with simulator outputs and verified mechanistic paths, and free of
contradictions with retrieved evidence?

\vspace{4pt}
Return ONLY valid JSON: \texttt{status} (\texttt{PASS}$|$\texttt{WARN}$|$\texttt{FAIL})
per check; \texttt{correction\_feedback} with concrete, targeted corrections for any
failed check. Failed checks trigger iterative refinement by the Mechanism Reasoning
Agent before final output is produced.
\end{tcolorbox}

\subsubsection{LLM Judge Prompt}
\begin{tcolorbox}[
  title={Prompt: Multi-Dimensional LLM Judge},
  colback=promptbg, colframe=promptframe,
  fonttitle=\small\bfseries, fontupper=\small,
  breakable, enhanced
]
\small
You are an expert scientific judge evaluating simulation-driven explanations produced
by different reasoning methods. Your evaluation must be rigorous, independent across
dimensions, and grounded in the provided simulation results. Score each explanation
on a \textbf{1--5 integer scale} across the following four dimensions.

\vspace{4pt}

\textbf{[Context Information]}\ \texttt{\{context\_summary\}}

\textbf{[Simulation Results]}\ \texttt{\{sim\_summaries\}}

\textbf{[Mechanism Graphs]}\ \texttt{\{mechanism\_graphs\}}

\textbf{[Explanations]}\ \textbf{A} (Causal-Copilot), \textbf{B} (Logic-LM),
\textbf{C} (GoT), \textbf{D} (MechSim): \texttt{\{explanations\}}

\vspace{4pt}
\textbf{1. Completeness} --- Are all $\mathcal{E}$-components
($\mathcal{I}, \mathcal{P}, \mathcal{Z}, \mathcal{C}$) present with sufficient depth?
\begin{itemize}[noitemsep, leftmargin=*]
  \item \textbf{5}: All four components present; each is substantive, specific, and
  mutually consistent.
  \item \textbf{4}: All components present; one is underdeveloped but does not
  undermine overall coherence.
  \item \textbf{3}: At least three components present; missing or shallow component
  noticeably limits the explanation.
  \item \textbf{2}: One or two components missing or largely superficial.
  \item \textbf{1}: Explanation is a generic summary with no meaningful
  $\mathcal{E}$-structure.
\end{itemize}

\vspace{4pt}
\textbf{2. Mechanistic Logic Consistency} --- Are reasoning steps logically consistent
with the simulator's mechanism structure and free of invalid transitions?
\begin{itemize}[noitemsep, leftmargin=*]
  \item \textbf{5}: Every reasoning step corresponds to a valid node-to-node
  transition in $\mathcal{G}_i$; no skipped states or spurious variables.
  \item \textbf{4}: Reasoning is largely graph-consistent; at most one minor
  transition error that does not affect the conclusion.
  \item \textbf{3}: Core propagation path is correct, but intermediate steps are
  incomplete or loosely connected to the graph.
  \item \textbf{2}: Multiple invalid transitions or variables introduced outside
  the mechanism graph.
  \item \textbf{1}: Reasoning ignores the mechanism structure entirely or contains
  contradictory transitions.
\end{itemize}

\vspace{4pt}
\textbf{3. Scientific Soundness} --- Are claims grounded in retrieved scientific
evidence and established domain knowledge?
\begin{itemize}[noitemsep, leftmargin=*]
  \item \textbf{5}: All major claims are supported by specific cited evidence;
  no assertions contradict established scientific consensus.
  \item \textbf{4}: Most claims are evidence-grounded; one claim is asserted
  without citation but is plausible and consistent with domain knowledge.
  \item \textbf{3}: Some claims are evidence-grounded; others are generic or
  imprecise; no direct contradictions with established science.
  \item \textbf{2}: Claims are mostly unsupported or rely on vague domain
  knowledge without specific grounding.
  \item \textbf{1}: Claims contradict established scientific evidence or are
  entirely fabricated without reference to retrieved literature.
\end{itemize}

\vspace{4pt}
\textbf{4. Faithfulness} --- Do claims accurately reflect simulation outputs
without hallucination or directional errors?
\begin{itemize}[noitemsep, leftmargin=*]
  \item \textbf{5}: All quantitative and directional claims are consistent with
  the simulation outputs; no hallucinated values or reversed effects.
  \item \textbf{4}: Claims are faithful overall; at most one minor imprecision
  that does not affect the recommendation.
  \item \textbf{3}: Most claims are directionally correct, but some lack
  quantitative grounding or overstate simulation findings.
  \item \textbf{2}: Noticeable hallucinations or directional errors in one or
  more claims.
  \item \textbf{1}: Claims systematically misrepresent simulation outputs or
  invert effect directions.
\end{itemize}

\vspace{4pt}
\textbf{Scoring instructions:} Evaluate each dimension independently for all
methods before moving to the next dimension. Do not anchor on the first explanation
read or on overall impression. Assign integer scores only. If two methods are
indistinguishable on a dimension, they may receive the same score.

\vspace{4pt}
Return ONLY valid JSON with the following structure: per-method integer scores
(1--5) for each of the four dimensions; \texttt{winner} (method with highest
average score); \texttt{ranking} (all methods ordered by average score,
descending); \texttt{score\_rationale} (one sentence per method per dimension
justifying the assigned score).
\end{tcolorbox}

% ═══════════════════════════════════════════════════════════════════════════════
\section{Additional Experimental Results}
\label{app:results}
% ═══════════════════════════════════════════════════════════════════════════════

\setcounter{table}{0}
\setcounter{figure}{0}

\subsection{Example Context Information}
\label{app:context_example}

Below is a representative context information block $\Omega$ extracted by the
Context Understanding Agent for a COVID-19 policy evaluation scenario. This
structured output is passed downstream to the Mechanism Modeling Agent,
Sensitivity Analysis Agent, and Mechanism Reasoning Agent.

\begin{tcolorbox}[
  title={Context Information $\Omega$: COVID-19 Policy Evaluation Scenario},
  colback=promptbg, colframe=mechblue,
  fonttitle=\small\bfseries, fontupper=\small,
  breakable, enhanced
]
\small

\noindent\textbf{Environment.}
Closed, well-mixed community of $N = 1{,}000{,}000$ individuals over a 180-day
horizon. Hospital capacity: 1,000 acute-care beds ($\approx 0.1\%$ of
population), representing a hard ceiling for simultaneous severe cases.
Key real-world factors include age-dependent severity, heterogeneous contact
patterns (households, workplaces, schools), and behavioral adaptation over time.

\vspace{6pt}
\noindent\textbf{Goal.}
Task: \textit{policy evaluation}. decision maker's question: determine why SIR, SEIRD, and
SEIR-HD yield divergent epidemic trajectories under the same inputs, and
identify the mechanisms driving those differences. Decision context:
public-health officials responsible for surge-capacity planning and intervention
timing.

\vspace{6pt}
\noindent\textbf{Key Entities.}

\noindent\textit{State variables}: $S,\ E,\ I,\ H,\ R,\ D$

\noindent\textit{Key parameters}:
\begin{itemize}[leftmargin=*, itemsep=1pt]
  \item $R_0 = 2.6$;\quad $\beta = 0.26$;\quad $\gamma = 0.10$;\quad
        $\sigma = 0.25$
  \item Hospital beds $= 1{,}000$;\quad simulation horizon $= 180$ days
  \item Initial infectious $I_0 = 100$
\end{itemize}

\noindent\textit{Intervention targets}: transmission rate $\beta$, contact
rate, vaccination rate, hospital capacity expansion.

\noindent\textit{Outcome metrics}: peak infection rate, total deaths,
hospital overload days, time to peak, cumulative cases.

\end{tcolorbox}

\subsection{Example Explanation Output}
\label{app:example_explanation}

We present a representative \mechsim explanation generated for the COVID-19 evaluate task.

\begin{tcolorbox}[
  title={MechSim Output: Example Structured Explanation
         $\mathcal{E} = (\mathcal{I}, \mathcal{P}, \mathcal{Z}, \mathcal{C})$},
  colback=promptbg, colframe=mechblue,
  fonttitle=\small\bfseries, fontupper=\small,
  breakable, enhanced
]
\small

\noindent\textbf{Context.}
US national COVID-19 outbreak ($N = 500{,}000$; $R_0 = 2.8$; hospital capacity
$= 3{,}000$ beds). Task: rank candidate intervention policies using three
simulators --- SIR, SEIRD, and SEIR-HD --- calibrated to observed weekly death
data. Evaluated policy: \textit{school closure +
work-from-home} (contact multiplier $\times 0.50$; cost $= 0.35$).

\vspace{6pt}
\noindent\textbf{1.~Output Interpretation ($\mathcal{I}$).}
All three simulators predict a reduction in peak daily infections under the
evaluated policy, but diverge substantially in magnitude and in the outcomes
they can represent. SIR projects $I_\text{max} \approx 180{,}000$ (day 55);
SEIRD projects $I_\text{max} \approx 155{,}000$ (day 63), with 8,900 cumulative
deaths versus SIR's 11,200. Critically, neither SIR nor SEIRD can represent
hospital occupancy, leaving the primary capacity-planning bottleneck unobserved.
SEIR-HD projects peak hospital occupancy of 12,400 beds --- exceeding the
3,000-bed threshold by $4\times$ --- sustained for 28 days, a signal invisible
to the other two simulators. This divergence has direct decision relevance: a
policy that appears effective under SIR or SEIRD may still produce system
overload under SEIR-HD.

\vspace{6pt}
\noindent\textbf{2.~Mechanism Reasoning Paths ($\mathcal{P}$).}

\noindent\textit{SIR} ($\mathcal{A}_i$: homogeneous mixing; constant $\beta$;
permanent immunity):
\[
S \xrightarrow{\beta(t)\downarrow} I \xrightarrow{\gamma} R
\]
School closure reduces the effective contact rate ($\beta: 0.26 \to 0.13$),
directly lowering the force of infection and suppressing the $S \to I$
transition. The absence of a latent compartment causes the peak to arrive
earlier and be overestimated relative to models with exposed states.

\noindent\textit{SEIRD} ($\mathcal{A}_i$: exposed latency $\sigma^{-1} = 4$
days; disease-induced mortality $\mu$):
\[
S \xrightarrow{\beta} E \xrightarrow{\sigma} I \xrightarrow{\gamma} R
\;\big/\; \xrightarrow{\mu} D
\]
The exposed compartment buffers the $S \to I$ flow, smoothing the epidemic
curve and shifting the peak by $+8$ days. Lower peak infections (13\% below
SIR) reflect latency damping. Mortality is modelled explicitly, but hospital
dynamics remain outside this simulator's scope.

\noindent\textit{SEIR-HD} ($\mathcal{A}_i$: hospitalisation branch; capacity
threshold; separate community vs.\ hospital CFR):
\[
S \xrightarrow{\beta} E \xrightarrow{\sigma} I \xrightarrow{\alpha} H
\xrightarrow{\gamma_H} R \;\big/\; \xrightarrow{\delta} D
\]
With hospitalisation rate $\alpha = 0.02$ and mean hospital stay
$\gamma_H^{-1} = 10$ days, the $I \to H$ branch sustains occupancy above
capacity for 28 days. Crucially, $\alpha$ is not modified by the school-closure
covariate, meaning adult severe cases accumulate at baseline rates even as
child-to-child transmission declines.

\vspace{6pt}
\noindent\textbf{3.~Supporting Evidence ($\mathcal{Z}$).}
\begin{itemize}[leftmargin=*, itemsep=3pt]
  \item \textbf{[Z1]} Cauchemez et al.\ (2008, \textit{Nature}).
  \textit{Estimating the impact of school closure on influenza transmission
  from sentinel data.} Reports 20--40\% reduction in transmission during school
  holidays in high-density settings. Supports the mechanistic effect of the
  $\beta \downarrow$ transition under school closure, though the upper bound of
  our assumed 50\% reduction warrants caution in adult-dominated contact networks.

  \item \textbf{[Z2]} Holmdahl \& Buckee (2020, \textit{New England Journal of
  Medicine}). \textit{Wrong but useful --- what COVID-19 epidemiologic models can
  and cannot tell us.} Demonstrates that hospital-explicit compartmental models
  are necessary for reliable capacity-planning forecasts, and that simpler
  compartmental models systematically underestimate overload risk. Directly
  supports the claim that SIR and SEIRD are structurally blind to the $I \to H$
  pathway.

  \item \textbf{[Z3]} Lauer et al.\ (2020, \textit{Annals of Internal Medicine}).
  \textit{The incubation period of coronavirus disease 2019 (COVID-19) from
  publicly reported confirmed cases.} Reports a median incubation period of 5.1
  days (95\% CI: 4.5--5.8 days). Validates SEIRD's latency assumption
  ($\sigma^{-1} = 4$ days) and supports the 8-day peak shift relative to SIR.

  \item \textbf{[Z4]} Ferguson et al.\ (2020, \textit{Imperial College London
  Report 9}). \textit{Impact of non-pharmaceutical interventions to reduce
  COVID-19 mortality and healthcare demand.} Projects that combined school
  closure and social distancing can suppress peak ICU demand by up to 67\%,
  but emphasizes that hospital surge capacity remains a binding constraint even
  under aggressive NPIs. Corroborates SEIR-HD's prediction of sustained overload
  despite policy implementation.
\end{itemize}

\vspace{6pt}
\noindent\textbf{4.~Claims ($\mathcal{C}$).}
\begin{enumerate}[leftmargin=*, itemsep=2pt]
  \item The \textit{homogeneous mixing} assumption shared by all three simulators
  uniformly scales all contact types, causing school closure to reduce adult
  workplace contacts in addition to child school contacts. This overestimates the
  policy effect for working-age adults.
  \hfill\textit{[Path: $\beta(t) \to S \to I$; Evidence: Z1]}

  \item SIR overestimates peak infections by 13\% relative to SEIRD because it
  lacks a latent compartment, causing exposed individuals to become immediately
  infectious. This compresses the epidemic curve and advances the peak by 8 days.
  \hfill\textit{[Path: $S \to I$ vs.\ $S \to E \to I$; Evidence: Z3]}

  \item SIR and SEIRD structurally cannot detect hospital overload: the $I \to H$
  branch is absent, making these simulators unreliable for capacity-planning
  decisions under any policy.
  \hfill\textit{[Evidence: Z2]}

  \item Under SEIR-HD, school closure does not reduce the hospitalisation rate
  $\alpha$, so severe adult cases accumulate at baseline rates. Even with
  infection suppression, hospital occupancy exceeds capacity by $4\times$ for
  28 days, indicating that transmission reduction alone is insufficient.
  \hfill\textit{[Path: $I \xrightarrow{\alpha} H$; $\alpha$ invariant to policy
  covariate; Evidence: Z4]}
\end{enumerate}

\vspace{6pt}
\noindent\textbf{5.~Decision Recommendation ($\mathcal{R}$).}
\textbf{Simulator selection}: Use SEIR-HD for forecasting and policy evaluation
in this scenario; it is the only simulator that exposes hospital overload risk,
which is the binding constraint under the current deployment context.
\textbf{Policy recommendation}: School closure should be implemented but is
insufficient alone. SEIR-HD predicts 28 days of capacity breach even under full
compliance; complementary measures targeting adult severe cases (e.g., targeted
elderly vaccination, hospital surge capacity expansion) are required.
\textbf{Uncertainty note}: The homogeneous mixing assumption shared by all
simulators may overstate the policy's effect on adult transmission; network-based
contact models should be considered if heterogeneous contact data are available.

\end{tcolorbox}

\subsection{Example Mechanism Graph}
\label{app:mechanism_graph}

Here shows the mechanism graph
$\mathcal{G}_i = (\mathcal{V}_i, \mathcal{E}_i, \mathcal{M}_i)$ automatically
extracted by the Mechanism Modeling Agent for the SEIR-HD simulator. Nodes
represent compartments ($\mathcal{V}_i$); directed edges represent mechanistic
transitions ($\mathcal{E}_i$), labelled with the governing rate; and graph-level
metadata ($\mathcal{M}_i$) includes simulator assumptions and execution statistics.

\begin{tcolorbox}[
  title={Graphviz DOT Code: SEIR-HD Mechanism Graph},
  colback=promptbg, colframe=promptframe,
  fonttitle=\small\bfseries, fontupper=\small,
  breakable, enhanced
]
\begin{verbatim}
digraph SEIR_HD {
    rankdir=LR;
    graph [
        label="SEIR-HD Mechanism Graph\n\
Assumptions:\n\
  A2: homogeneous mixing\n\
  A3: latent (incubation) period\n\
  A4: hospitalisation branch\n\
  A5: hospital CFR (recovery vs. death)\n\
  A6: permanent immunity post-recovery\n\
  A7: deterministic ODE dynamics\n\
--------------------------------------------\n\
Execution Statistics (calibrated run):\n\
  Peak Infection Rate : 0.0128\n\
  Total Deaths        : 2764\n\
  Hospital Overload   : 69 days",
        labelloc=b,
        fontsize=11,
        fontname="Helvetica"
    ];

    /* -- Node declarations -- */
    node [shape=circle, style=filled,
          fontname="Helvetica", fontsize=12];

    S [label="S\nSusceptible",  fillcolor="#AED6F1"];
    E [label="E\nExposed",      fillcolor="#A9DFBF"];
    I [label="I\nInfectious",   fillcolor="#F9E79F"];
    H [label="H\nHospitalised", fillcolor="#F0B27A"];
    R [label="R\nRecovered",    fillcolor="#D7BDE2"];
    D [label="D\nDeceased",     fillcolor="#F1948A"];

    /* -- Mechanistic edges -- */
    edge [fontname="Helvetica", fontsize=10];

    S -> E [label="beta*S*I/N  [A2]",
            color="#2E86C1", fontcolor="#2E86C1"];

    E -> I [label="sigma (1/incubation)  [A3]",
            color="#1E8449", fontcolor="#1E8449"];

    I -> R [label="gamma (community recovery)  [A4]",
            color="#7D3C98", fontcolor="#7D3C98"];

    I -> H [label="theta (hospitalisation rate)  [A4]",
            color="#CA6F1E", fontcolor="#CA6F1E",
            style=bold];

    H -> R [label="gamma_H (hospital recovery)  [A5]",
            color="#7D3C98", fontcolor="#7D3C98",
            style=dashed];

    H -> D [label="mu_H (hospital CFR)  [A5]",
            color="#C0392B", fontcolor="#C0392B",
            style=bold];

    /* -- Primary propagation path (S->E->I->H->D) -- */
    {rank=same; S; E; I; H; D}
}
\end{verbatim}
\end{tcolorbox}

\subsection{Simulation Result}
\label{app:sim_results}

\paragraph{COVID-19 Forecast Task.}
Table~\ref{tab:covid_forecast} reports simulator performance on held-out US COVID-19 data. The task requires selecting the best simulator prior to the forecast cutoff using training-period evidence alone.

\begin{table}[H]
\centering
\caption{COVID-19 simulator forecast performance on held-out period (US national). Lower WIS, CRPS, nRMSE are better; higher coverage is better.}
\label{tab:covid_forecast}
\small
\begin{tabular}{lcccc}
\toprule
\textbf{Simulator} & \textbf{Mean WIS} & \textbf{Mean CRPS} & \textbf{nRMSE} & \textbf{Cov.\ (90\%)}\\
\midrule
SIR (Dynamic)       & 87,312 & 14,821 & 1.031 & 0.712\\
SEIRD (Dynamic)     & 79,450 & 13,642 & 0.924 & 0.764\\
SEIR-HD             & 68,234 & 11,508 & 0.847 & 0.829\\
SEIQRD              & 73,891 & 12,201 & 0.889 & 0.801\\
SEAIHRD             & 71,105 & 11,944 & 0.872 & 0.815\\
SIRV                & 80,117 & 13,299 & 0.951 & 0.758\\
\midrule
ParticleFilter-SEIRH (ref.) & 73,485 & 12,354 & 0.964 & 0.890\\
ARIMA (ref.)        & 52,688 &  7,216 & 0.438 & 0.829\\
Constant (ref.)     & 115,269 & 10,978 & 0.484 & 0.000\\
\bottomrule
\end{tabular}
\end{table}

\paragraph{Supply Chain Policy Selection Task.}
Table~\ref{tab:sc_scenario2} reports simulator-predicted harm scores across six
candidate policies for a representative scenario (Scenario~2). The oracle
identifies Policy~4 (Expedited Replenishment) as optimal.

\begin{table}[H]
\centering
\caption{Simulator-predicted harm across six candidate policies (Scenario~2,
Supply Chain). Values in \$k. \textbf{Bold} = oracle-optimal policy (P4).}
\label{tab:sc_scenario2}
\small
\setlength{\tabcolsep}{10pt}
\begin{tabular}{l r r r r r r}
\toprule
\textbf{Simulator}
  & \textbf{P1} & \textbf{P2} & \textbf{P3}
  & \textbf{P4} & \textbf{P5} & \textbf{P6} \\
\midrule
Oracle
  & 74.0 & 102.7 & 96.2 & \textbf{58.1} & 96.5 & 99.7 \\
\midrule
Base Stock
  & \phantom{00}0.6 & 135.1 & 225.0 & 113.1 & \phantom{0}81.0 & 315.0 \\
$(s,S)$ Inventory
  & \phantom{00}0.9 & 135.9 & 225.6 & 113.4 & \phantom{0}81.9 & 315.9 \\
Multi-Echelon
  & \phantom{0}12.1 & \phantom{0}12.5 & \phantom{0}12.3
  & \phantom{0}12.2 & \phantom{0}12.2 & \phantom{0}12.7 \\
Beer Game
  & \phantom{0}21.1 & 156.1 & 246.1 & 133.6 & 102.1 & 336.1 \\
\bottomrule
\end{tabular}
\end{table}

\subsection{Sensitivity Analysis Results}
\label{app:sensitivity_results}

Table~\ref{tab:sensitivity_covid} shows parametric sensitivity scores for the COVID-19 default scenario. Scores measure the interventional effect $\mathbb{E}[y \mid \mathrm{do}(\phi{=}\phi_2)] - \mathbb{E}[y \mid \mathrm{do}(\phi{=}\phi_1)]$. Table~\ref{tab:structural_sensitivity} summarises structural sensitivity results from assumption-level interventions.

\begin{table}[H]
\centering
\caption{Parametric sensitivity analysis: COVID-19 evaluate task, outcome = peak infection rate. Rank 1 is most influential.}
\label{tab:sensitivity_covid}
\small
\begin{tabular}{lcccp{4cm}}
\toprule
\textbf{Parameter} & \textbf{Score} & \textbf{Rank} & \textbf{Direction} & \textbf{Interpretation}\\
\midrule
Transmission rate ($R_0$)     & 5.57 & 1 & Positive  & Higher $R_0 \to$ higher, earlier peak\\
Contact rate                  & 2.35 & 2 & Positive  & Higher contact rate $\to$ faster spread\\
Behavioral reduction ($\alpha$) & 0.85 & 3 & Negative & Stronger NPI $\to$ lower peak\\
Recovery rate ($\gamma$)      & 0.81 & 4 & Negative  & Faster recovery $\to$ lower, later peak\\
\bottomrule
\end{tabular}
\end{table}

\begin{table}[H]
\centering
\caption{Structural sensitivity analysis: impact of modifying key modeling assumptions.}
\label{tab:structural_sensitivity}
\small
\begin{tabular}{llccc}
\toprule
\textbf{Assumption} & \textbf{Variant A} & \textbf{Variant B} & \textbf{$\Delta$ Peak} & \textbf{$\Delta$ Deaths}\\
\midrule
Mixing pattern & Homogeneous & Network-based  & $+14.9\%$ & $+8{,}855$\\
Immunity       & Permanent   & Waning (180d)  & $+22.3\%$ & $+15{,}240$\\
Severity dist. & Mild-dominant & Severe-dominant & $+8.1\%$ & $+31{,}100$\\
\bottomrule
\end{tabular}
\end{table}

\subsection{Ablation Study Results}
\label{app:ablation}
\begin{figure}[t]
    \centering
    \includegraphics[width=0.8\linewidth]{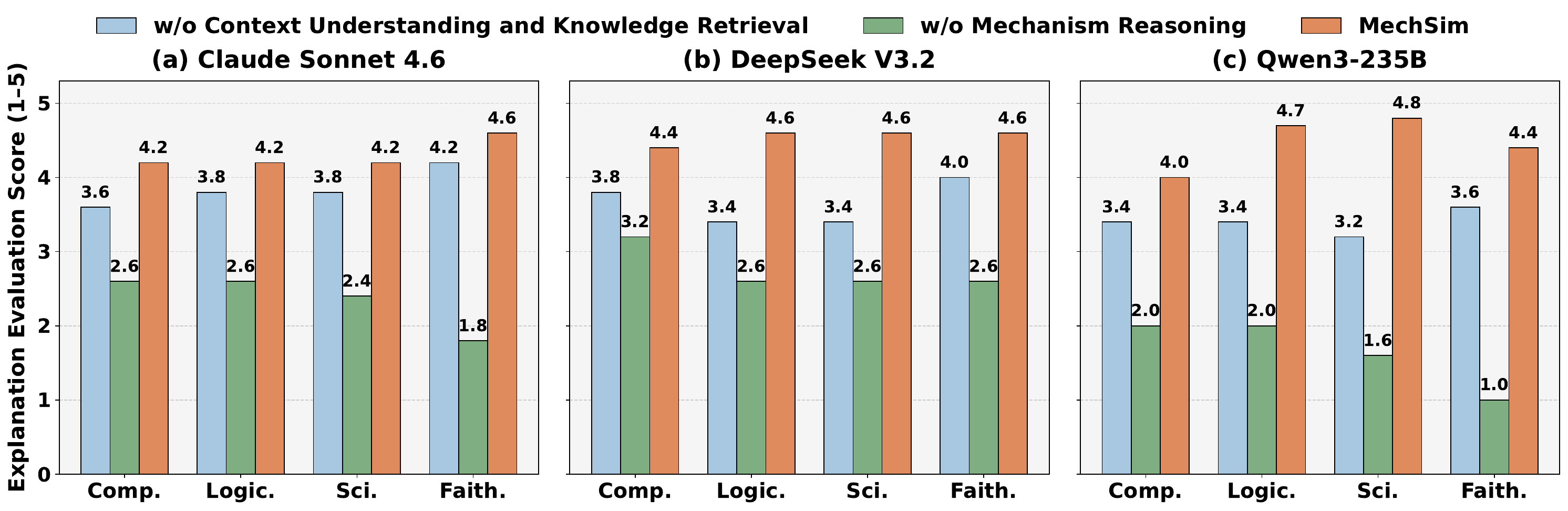}
    \caption{Selected COVID-19 ablation results for the explanation evaluation task across three LLM backbones. ``w/o'' denotes removal of the module or module group named in the legend.}
    \label{fig:ablation_study}
\end{figure}

Figure~\ref{fig:ablation_study} shows selected COVID-19 ablation results. Removing mechanism reasoning causes the largest degradation, with relative score reductions of around $50\%$ on completeness, mechanism-logic consistency, scientific soundness, and faithfulness. This shows that explicit mechanism reasoning contributes to the framework's explanation gains. Removing context understanding and knowledge retrieval has a smaller but consistent effect, with relative reductions of around $18\%$ on the same four criteria. Overall, the ablation results suggest that the framework's components work together to support high-quality explanations, with mechanism reasoning playing a central role. In addition, the degradation is larger for less powerful LLMs, suggesting that the framework's structured reasoning may be especially beneficial for weaker LLMs. 

\subsection{Comprehensive Policy Selection Task Result}
\label{com_policy_selection}
We put the Comprehensive Policy Selection Task Result with standard deviation here in \ref{tab:app_policy_selection}.
\begin{table}[t]
\centering
\small
\setlength{\tabcolsep}{6pt}
\renewcommand{\arraystretch}{1.2}
\caption{Policy selection results (Precision@$k$ and Recall@$k$) across three domains and three LLM backbones. \textbf{Bold} = best per metric within each domain/LLM group. ``GoT'' denotes Graph of Thoughts.}
\label{tab:app_policy_selection}
\begin{tabular}{l l l rrrr}
\toprule
\textbf{Domain} & \textbf{LLM} & \textbf{Method} & \textbf{P@3} & \textbf{P@5} & \textbf{R@3} & \textbf{R@5} \\
\midrule

\multirow{12}{*}{COVID-19}
 & \multirow{4}{*}{Claude}
   & Causal-Copilot  & $0.667_{\pm.000}$ & $0.720_{\pm.103}$ & $0.400_{\pm.024}$ & $0.701_{\pm.105}$ \\
 & & Logic-LM & $0.633_{\pm.105}$ & $0.620_{\pm.114}$ & $0.380_{\pm.063}$ & $0.645_{\pm.138}$ \\
 & & GoT      & $0.633_{\pm.105}$ & $0.600_{\pm.133}$ & $0.380_{\pm.063}$ & $0.624_{\pm.133}$ \\
 & & MechSim  & $\mathbf{0.823}_{\pm.161}$ & $\mathbf{0.714}_{\pm.105}$ & $\mathbf{0.540}_{\pm.097}$ & $\mathbf{0.720}_{\pm.103}$ \\
\cmidrule(lr){2-7}
 & \multirow{4}{*}{DeepSeek}
   & Causal-Copilot  & $0.633_{\pm.105}$ & $0.500_{\pm.105}$ & $0.380_{\pm.063}$ & $0.700_{\pm.105}$ \\
 & & Logic-LM & $0.600_{\pm.141}$ & $0.640_{\pm.084}$ & $0.360_{\pm.084}$ & $0.640_{\pm.084}$ \\
 & & GoT      & $0.667_{\pm.000}$ & $0.520_{\pm.103}$ & $0.423_{\pm.036}$ & $0.520_{\pm.103}$ \\
 & & MechSim  & $\mathbf{0.833}_{\pm.176}$ & $\mathbf{0.720}_{\pm.103}$ & $\mathbf{0.500}_{\pm.105}$ & $\mathbf{0.720}_{\pm.103}$ \\
\cmidrule(lr){2-7}
 & \multirow{4}{*}{Qwen}
   & Causal-Copilot  & $0.667_{\pm.000}$ & $0.680_{\pm.103}$ & $0.640_{\pm.097}$ & $0.580_{\pm.132}$ \\
 & & Logic-LM & $0.700_{\pm.105}$ & $0.600_{\pm.094}$ & $0.420_{\pm.105}$ & $0.667_{\pm.086}$ \\
 & & GoT      & $0.804_{\pm.142}$ & $0.520_{\pm.103}$ & $0.650_{\pm.105}$ & $0.721_{\pm.124}$ \\
 & & MechSim  & $\mathbf{0.833}_{\pm.176}$ & $\mathbf{0.700}_{\pm.105}$ & $\mathbf{0.760}_{\pm.084}$ & $\mathbf{0.766}_{\pm.132}$ \\

\midrule

\multirow{12}{*}{\shortstack[l]{Supply\\Chain}}
 & \multirow{4}{*}{Claude}
   & Causal-Copilot  & $0.783_{\pm.163}$ & $0.810_{\pm.045}$ & $0.470_{\pm.098}$ & $0.543_{\pm.231}$ \\
 & & Logic-LM & $0.817_{\pm.170}$ & $0.830_{\pm.073}$ & $0.490_{\pm.102}$ & $0.350_{\pm.408}$ \\
 & & GoT      & $0.783_{\pm.163}$ & $0.812_{\pm.089}$ & $0.470_{\pm.098}$ & $0.544_{\pm.357}$ \\
 & & MechSim  & $\mathbf{0.900}_{\pm.157}$ & $\mathbf{0.840}_{\pm.082}$ & $\mathbf{0.540}_{\pm.094}$ & $\mathbf{0.569}_{\pm.183}$ \\
\cmidrule(lr){2-7}
 & \multirow{4}{*}{DeepSeek}
   & Causal-Copilot  & $0.833_{\pm.171}$ & $0.810_{\pm.045}$ & $0.500_{\pm.103}$ & $0.810_{\pm.045}$ \\
 & & Logic-LM & $0.867_{\pm.168}$ & $0.850_{\pm.089}$ & $0.520_{\pm.101}$ & $0.850_{\pm.089}$ \\
 & & GoT      & $0.833_{\pm.171}$ & $0.810_{\pm.045}$ & $0.500_{\pm.103}$ & $0.810_{\pm.045}$ \\
 & & MechSim  & $\mathbf{0.967}_{\pm.103}$ & $\mathbf{0.860}_{\pm.094}$ & $\mathbf{0.580}_{\pm.062}$ & $\mathbf{0.860}_{\pm.094}$ \\
\cmidrule(lr){2-7}
 & \multirow{4}{*}{Qwen}
   & Causal-Copilot  & $0.856_{\pm.168}$ & $0.833_{\pm.076}$ & $0.513_{\pm.101}$ & $0.488_{\pm.170}$ \\
 & & Logic-LM & $0.867_{\pm.166}$ & $\mathbf{0.856}_{\pm.186}$ & $0.520_{\pm.100}$ & $0.490_{\pm.241}$ \\
 & & GoT      & $0.845_{\pm.169}$ & $0.840_{\pm.081}$ & $0.507_{\pm.102}$ & $0.501_{\pm.188}$ \\
 & & MechSim  & $\mathbf{0.911}_{\pm.150}$ & $0.842_{\pm.046}$ & $\mathbf{0.547}_{\pm.090}$ & $\mathbf{0.524}_{\pm.192}$ \\

\midrule

\multirow{12}{*}{Measles}
 & \multirow{4}{*}{Claude}
   & Causal-Copilot  & $0.556_{\pm.248}$ & $0.733_{\pm.149}$ & $0.363_{\pm.124}$ & $0.742_{\pm.109}$ \\
 & & Logic-LM & $\mathbf{0.722}_{\pm.299}$ & $0.767_{\pm.180}$ & $0.367_{\pm.143}$ & $\mathbf{0.833}_{\pm.180}$ \\
 & & GoT      & $0.333_{\pm.038}$ & $0.631_{\pm.187}$ & $0.293_{\pm.902}$ & $0.657_{\pm.103}$ \\
 & & MechSim  & $0.611_{\pm.299}$ & $\mathbf{0.833}_{\pm.180}$ & $\mathbf{0.433}_{\pm.180}$ & $0.767_{\pm.129}$ \\
\cmidrule(lr){2-7}
 & \multirow{4}{*}{DeepSeek}
   & Causal-Copilot  & $0.611_{\pm.299}$ & $0.767_{\pm.180}$ & $0.367_{\pm.146}$ & $0.767_{\pm.197}$ \\
 & & Logic-LM & $0.722_{\pm.404}$ & $0.667_{\pm.249}$ & $\mathbf{0.433}_{\pm.243}$ & $0.667_{\pm.249}$ \\
 & & GoT      & $0.579_{\pm.196}$ & $0.425_{\pm.127}$ & $0.428_{\pm.579}$ & $0.648_{\pm.209}$ \\
 & & MechSim  & $\mathbf{0.767}_{\pm.180}$ & $\mathbf{0.767}_{\pm.180}$ & $0.379_{\pm.180}$ & $\mathbf{0.767}_{\pm.180}$ \\
\cmidrule(lr){2-7}
 & \multirow{4}{*}{Qwen}
   & Causal-Copilot  & $0.667_{\pm.272}$ & $0.767_{\pm.180}$ & $0.400_{\pm.163}$ & $0.767_{\pm.180}$ \\
 & & Logic-LM & $0.722_{\pm.299}$ & $0.733_{\pm.275}$ & $0.467_{\pm.221}$ & $0.733_{\pm.275}$ \\
 & & GoT      & $0.333_{\pm.138}$ & $0.648_{\pm.128}$ & $\mathbf{0.583}_{\pm.256}$ & $0.529_{\pm.108}$ \\
 & & MechSim  & $\mathbf{0.778}_{\pm.369}$ & $\mathbf{0.833}_{\pm.180}$ & $0.433_{\pm.139}$ & $\mathbf{0.833}_{\pm.153}$ \\

\bottomrule
\end{tabular}
\end{table}

\subsection{Comprehensive Simulator Selection Task Result}
\label{app_simulator_selection}
We put the Comprehensive Simulator Selection Task Result with standard deviation here \ref{tab:simulator_selection_full}.

\begin{table*}[t]
\centering
\small
\setlength{\tabcolsep}{6pt}
\renewcommand{\arraystretch}{1.25}
\caption{Full simulator selection results with standard deviations (Top-$k$ regret, $k\in\{1,3\}$; \emph{lower is better}). Values are reported as mean$\pm$std. \textbf{Bold} denotes the best (lowest) mean within each (LLM, Domain, Metric) group.}
\label{tab:simulator_selection_full}
\begin{tabular*}{\textwidth}{@{\extracolsep{\fill}} l l l cccc @{}}
\toprule
\textbf{LLM} & \textbf{Domain} & \textbf{Metric} & \textbf{Causal-Copilot} & \textbf{Logic-LM} & \textbf{GoT} & \textbf{MechSim (Ours)} \\
\midrule

\multirow{6}{*}{\rotatebox[origin=c]{90}{\textbf{Claude Sonnet 4.6}}}
 & \multirow{2}{*}{COVID-19}
   & Top-1 Regret & $2.590{\pm}2.340$ & $3.910{\pm}3.180$ & $3.800{\pm}3.000$ & $\mathbf{2.040{\pm}2.000}$ \\
 & & Top-3 Regret & $2.600{\pm}1.760$ & $2.410{\pm}1.730$ & $2.530{\pm}1.690$ & $\mathbf{2.290{\pm}1.750}$ \\
\cmidrule(l){2-7}
 & \multirow{2}{*}{\makecell[l]{Supply\\Chain}}
   & Top-1 Regret & $6.170{\pm}3.480$ & $5.420{\pm}4.020$ & $5.920{\pm}0.370$ & $\mathbf{4.250{\pm}4.390}$ \\
 & & Top-3 Regret & $5.350{\pm}3.650$ & $4.380{\pm}4.310$ & $5.140{\pm}0.391$ & $\mathbf{3.470{\pm}4.140}$ \\
\cmidrule(l){2-7}
 & \multirow{2}{*}{Measles}
   & Top-1 Regret & $2.880{\pm}3.900$ & $2.350{\pm}3.090$ & $1.930{\pm}3.210$ & $\mathbf{1.690{\pm}1.140}$ \\
 & & Top-3 Regret & $3.400{\pm}4.530$ & $3.210{\pm}2.060$ & $2.790{\pm}2.830$ & $\mathbf{2.200{\pm}3.530}$ \\
\midrule

\multirow{6}{*}{\rotatebox[origin=c]{90}{\textbf{DeepSeek V3}}}
 & \multirow{2}{*}{COVID-19}
   & Top-1 Regret & $2.380{\pm}2.690$ & $2.700{\pm}3.020$ & $2.470{\pm}2.730$ & $\mathbf{2.160{\pm}2.620}$ \\
 & & Top-3 Regret & $2.100{\pm}1.800$ & $2.140{\pm}1.810$ & $2.180{\pm}1.880$ & $\mathbf{2.110{\pm}1.790}$ \\
\cmidrule(l){2-7}
 & \multirow{2}{*}{\makecell[l]{Supply\\Chain}}
   & Top-1 Regret & $0.617{\pm}0.348$ & $0.542{\pm}0.402$ & $0.592{\pm}0.370$ & $\mathbf{0.467{\pm}0.407}$ \\
 & & Top-3 Regret & $0.535{\pm}0.365$ & $0.438{\pm}0.431$ & $0.514{\pm}0.391$ & $\mathbf{0.389{\pm}0.393}$ \\
\cmidrule(l){2-7}
 & \multirow{2}{*}{Measles}
   & Top-1 Regret & $\mathbf{0.548{\pm}1.217}$ & $5.940{\pm}8.936$ & $2.486{\pm}4.204$ & $0.702{\pm}1.181$ \\
 & & Top-3 Regret & $2.994{\pm}4.146$ & $3.261{\pm}4.235$ & $2.764{\pm}3.469$ & $\mathbf{2.278{\pm}2.877}$ \\
\midrule

\multirow{6}{*}{\rotatebox[origin=c]{90}{\textbf{Qwen3-235B}}}
 & \multirow{2}{*}{COVID-19}
   & Top-1 Regret & $1.790{\pm}0.850$ & $1.780{\pm}0.870$ & $1.760{\pm}0.860$ & $\mathbf{1.650{\pm}0.880}$ \\
 & & Top-3 Regret & $1.700{\pm}0.750$ & $1.850{\pm}0.910$ & $1.700{\pm}0.750$ & $\mathbf{1.680{\pm}0.830}$ \\
\cmidrule(l){2-7}
 & \multirow{2}{*}{\makecell[l]{Supply\\Chain}}
   & Top-1 Regret & $0.593{\pm}0.334$ & $0.621{\pm}0.361$ & $0.591{\pm}0.338$ & $\mathbf{0.240{\pm}0.306}$ \\
 & & Top-3 Regret & $0.556{\pm}0.162$ & $0.514{\pm}0.190$ & $0.556{\pm}0.162$ & $\mathbf{0.513{\pm}0.216}$ \\
\cmidrule(l){2-7}
 & \multirow{2}{*}{Measles}
   & Top-1 Regret & $1.870{\pm}4.108$ & $2.693{\pm}2.999$ & $3.036{\pm}5.321$ & $\mathbf{1.128{\pm}2.008}$ \\
 & & Top-3 Regret & $2.549{\pm}3.183$ & $3.172{\pm}4.287$ & $2.698{\pm}3.427$ & $\mathbf{2.488{\pm}4.187}$ \\

\bottomrule
\end{tabular*}
\end{table*}

\end{document}